\documentclass[letterpaper,journal]{IEEEtran}
\usepackage{amsmath,amsfonts}
\usepackage{algorithmic}
\usepackage{array}
\usepackage[caption=false,font=normalsize,labelfont=sf,textfont=sf]{subfig}
\usepackage{textcomp}
\usepackage{stfloats}
\usepackage{url}
\usepackage{verbatim}
\usepackage{graphicx}
\def\BibTeX{{\rm B\kern-.05em{\sc i\kern-.025em b}\kern-.08em
    T\kern-.1667em\lower.7ex\hbox{E}\kern-.125emX}}
\usepackage{balance}

\usepackage{algorithm}
\usepackage{algorithmic}
\usepackage{amssymb}
\usepackage{amsmath}
\usepackage{microtype}
\usepackage{booktabs}
\usepackage{threeparttable}
\usepackage{multirow}
\usepackage{makecell}
\usepackage{colortbl}
\usepackage[table,xcdraw]{xcolor}
\definecolor{bestred}{HTML}{B22222}
\definecolor{secondblue}{HTML}{1F5AA6}
\newcommand{\bestresult}[1]{\textcolor{bestred}{\textbf{#1}}}
\newcommand{\secondresult}[1]{\textcolor{secondblue}{\underline{#1}}}
\usepackage{marvosym}

\begin{document}
\title{BEAT: Balanced Frequency Adaptive Tuning for Long-Term Time-Series Forecasting}
\author{Zhixuan~Li, Naipeng~Chen, Seonghwa~Choi, Sanghoon~Lee,~\IEEEmembership{Fellow,~IEEE},
and Weisi~Lin,~\IEEEmembership{Fellow,~IEEE}%
\thanks{This research was supported by the Ministry of Education of Singapore Tier3 grant MOET32022-0006. This work was supported in part by Culture, Sports and Tourism R\&DProgram through Korea Creative Content Agency grant funded by the Ministry of Culture, Sports and Tourism in 2024 under Grant RS-2024-00398413. (Corresponding author: Weisi Lin.)}%
\thanks{Zhixuan Li, Naipeng Chen, and Weisi Lin are with the College of Computing and Data Science, Nanyang Technological University, Singapore (e-mail: wslin@ntu.edu.sg).}%
\thanks{Seonghwa Choi and Sanghoon Lee are with the Department of Electrical and Electronic Engineering, Yonsei University, Korea.}}

\markboth{BEAT: Balanced Frequency Adaptive Tuning}%
{BEAT: Balanced Frequency Adaptive Tuning}

\maketitle

\begin{abstract}
Long-term time-series forecasting supports a wide range of applications, including weather prediction and electricity demand planning. Frequency-domain methods address this task by decomposing observations into components that describe temporal variations at different scales. However, separate representations do not by themselves provide an explicit mechanism for adjusting the training emphasis across components. Under a shared forecasting objective, the frequency-specific networks can retain different levels of coefficient prediction error, motivating an error-dependent adjustment to their gradients. To this end, we propose BEAT (Balanced frEquency Adaptive Tuning), a framework that combines frequency-specific error monitoring with adaptive gradient modulation. We design a Frequency-Specific Monitor that compares predicted and target wavelet coefficients in a common normalized space and expresses each discrepancy relative to a reference error computed from the detail components. We further introduce a Dynamical Gradient Balancer that converts these ratios into positive, bounded coefficients. Components with higher relative errors receive larger gradient weights, whereas those with lower relative errors receive smaller weights. A shared modulation-strength parameter controls the departure from unmodulated training, and the monitoring and balancing operations are used only during training. Experiments on seven real-world datasets show that BEAT achieves competitive performance against state-of-the-art forecasting methods.
\end{abstract}

\begin{IEEEkeywords}
Adaptive learning, Gradient modulation, Long-term time-series forecasting, Multiscale learning, Wavelet decomposition.
\end{IEEEkeywords}

\section{Introduction}
\label{sec:introduction}

\begin{figure*}[!t]
\centering
\includegraphics[width=\textwidth]{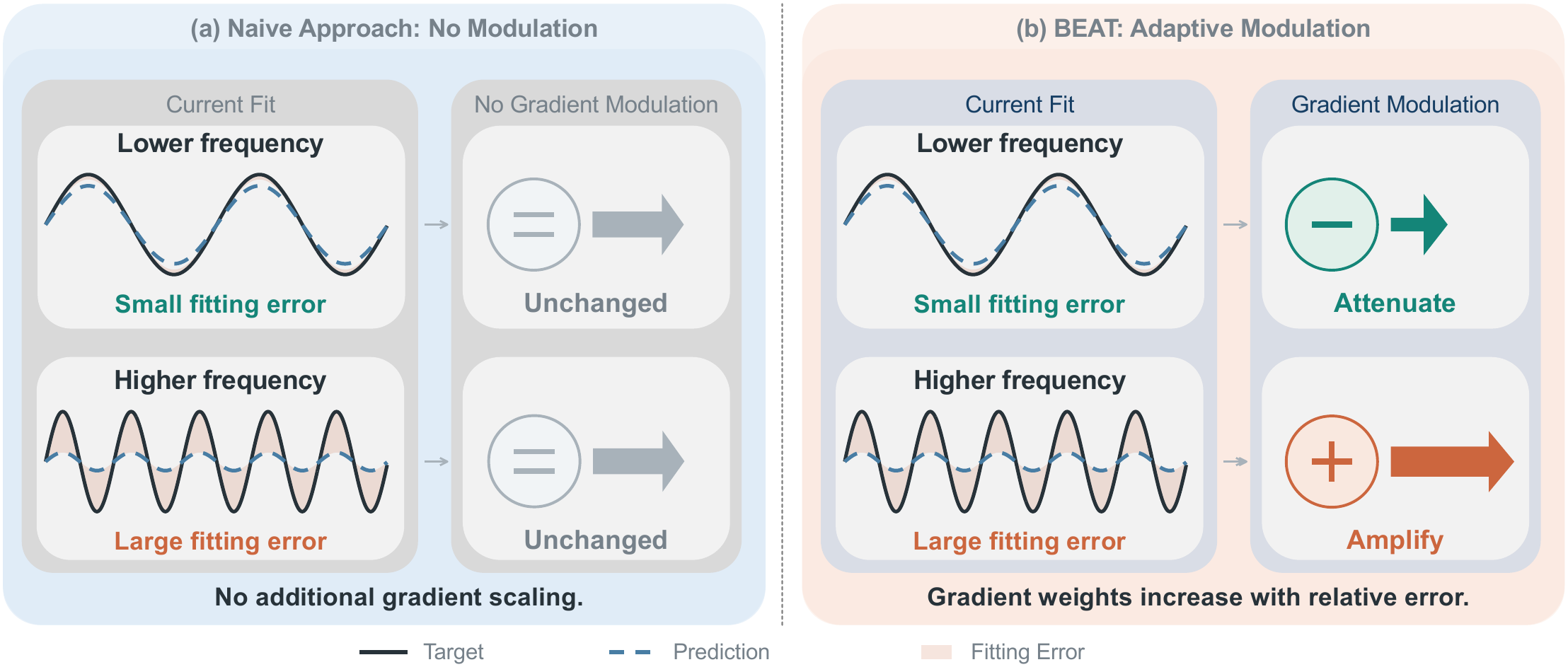}
\caption{An illustrative case of error-dependent gradient modulation. The two panels show identical target and predicted coefficient series in the same normalized space, rather than experimental results before and after training. (a) The naive approach applies no additional gradient scaling. (b) BEAT attenuates the gradient of the illustrated lower-frequency branch, whose fitting error is below the detail-based reference, and amplifies that of the illustrated higher-frequency branch, whose error is above the reference. The two displayed branches are examples from a multilevel decomposition. Their error ordering is case-specific, not a fixed preference for low or high frequencies. Arrow sizes indicate scaling relative to each branch's own unmodulated gradient, not equal gradient magnitudes across branches.}
\label{fig:motivation}
\end{figure*}

Long-term time-series forecasting supports applications such as weather prediction~\cite{zhang2022solar}, electric load forecasting~\cite{zhou2021informer,gasparin2022deep}, traffic flow analysis~\cite{jin2021trafficbert}, and financial market modeling~\cite{lai2018modeling,tang2022survey}. These tasks require models to use historical observations to predict temporal variations over extended horizons. Forecasting approaches address this challenge with CNN-based~\cite{liu2022scinet,wang2023micn,luo2024moderntcn}, linear- and MLP-based~\cite{oreshkin2019nbeats,zeng2023transformers,wang2024timemixer}, Transformer-based~\cite{wu2021autoformer,liu2021pyraformer,liu2023itransformer}, and LLM-based~\cite{jin2024time,jia2024gpt4mts} architectures. Alongside architectural development, frequency-domain methods~\cite{wu2021autoformer,wu2023timesnet,zhou2022fedformer} provide a complementary representation of temporal variation, allowing forecasting models to distinguish components associated with different scales. Fredformer~\cite{piao2024fredformer} further addresses frequency bias caused by disproportionate attention to high-energy components, highlighting the importance of learning across frequency bands.

Wavelet decomposition makes this separation explicit through approximation and detail coefficient series. The approximation represents slowly varying behavior, while the details describe variations at finer temporal resolutions. WaveToken~\cite{masserano2025wavetoken} uses quantized wavelet coefficients as tokens for autoregressive forecasting, illustrating their value as a predictive representation. WPMixer~\cite{murad2024wpmixer} combines multi-level wavelet decomposition, patching, and mixing to predict these series with separate frequency-specific networks. Their outputs are subsequently reconstructed into a complete forecast. This organization allows each network to model one component while retaining a shared forecasting objective. However, separating the representations does not by itself specify how the training emphasis should respond to differences in the fitting errors of those components.

Training objectives offer another way to use transformed representations. FreDF~\cite{wang2025fredf} aligns forecasts and targets in the frequency domain to address label correlation. Time-o1~\cite{wang2025timeo1} emphasizes significant decorrelated label components, while OLMA~\cite{shi2025olma} combines Fourier- and wavelet-domain supervision to alleviate frequency bias. These approaches motivate attention to the training process beyond the forward architecture. BEAT focuses on a different intervention, using the relative coefficient errors to modulate the gradients of frequency-specific prediction branches.

The distinction matters because a reconstructed forecast combines predictions from several branches, each of which can retain a different coefficient prediction error. Back-propagation of the overall forecasting loss already produces branch-dependent gradients through the reconstruction path. Nevertheless, these gradients do not include an explicit multiplier based on how each branch's coefficient error compares with a shared reference. An aggregate loss value also does not identify the distribution of fitting errors across the coefficient series. This motivates a complementary training signal that monitors the components directly, while preserving the forecasting objective that connects all branches to the final prediction.

Consider the illustrative training case in Figure~\ref{fig:motivation}, in which the lower-frequency branch fits its coefficient target more closely than the higher-frequency branch. The naive approach in Figure~\ref{fig:motivation}(a) applies no additional gradient scaling to account for this difference. A natural response is to assign less gradient weight to the better-fitted branch and more to the less well-fitted one, as illustrated in Figure~\ref{fig:motivation}(b). Such weighting should follow the branches' current relative fitting errors and adapt as training progresses, rather than remain tied to a particular frequency band. The intuition is that giving relatively poorly fitted components greater training emphasis may help reduce fitting imbalances across branches and improve the final forecast reconstructed from their outputs.

To this end, we propose BEAT (Balanced frEquency Adaptive Tuning), a framework that couples frequency-specific error monitoring with bounded gradient modulation. Inspired by~\cite{peng2022balanced}, BEAT is designed to formulate the monitoring signal using wavelet coefficient prediction errors. 
Specifically, we design a Frequency-Specific Monitor (FSM) to make the fitting state of each component available to the modulation rule. FSM first maps the ground-truth target to the same normalized space as the predictions using statistics from the historical input. It then applies the same wavelet configuration as the prediction branches and compares corresponding predicted and target coefficients. Each discrepancy is expressed relative to the mean discrepancy of the detail components. The approximation branch is assessed against the same reference but does not contribute to its calculation. These measurements describe the current relative fitting errors and are refreshed for each mini-batch, without requiring a stored trajectory of past losses.

We further introduce a Dynamical Gradient Balancer (DGB) that maps these measurements to positive, bounded gradient coefficients. Branches above the reference receive amplification, whereas those below it receive attenuation. A shared, tunable strength parameter controls the extent of the adjustment. Bounding the multiplier limits the amplification attributable to modulation, while keeping it positive preserves the direction of the original branch gradient. The overall forecasting loss remains responsible for computing those gradients, and FSM determines the information used to weight them. Both monitoring and modulation are confined to training, leaving inference to the frequency-specific predictors and wavelet reconstruction.

The reported benchmark comparison covers seven real-world datasets and multiple forecasting horizons. It shows lower horizon-averaged MSE and MAE than the WPMixer backbone on all seven datasets, although the best-performing method varies across datasets. The reported ablations on Weather and ETTh1 further examine the combined monitoring and balancing components, wavelet configurations, discrepancy measures, and modulation designs. These comparisons assess forecasting accuracy and the design choices of frequency-specific training.

The main contributions are summarized as follows.
\begin{itemize}
    \item We formulate an error-dependent training strategy for wavelet-based forecasting, using relative coefficient prediction errors to adjust the training emphasis across frequency-specific branches.
    \item We develop BEAT with two complementary components, FSM and DGB. FSM monitors relative fitting errors, and DGB translates them into positive, bounded gradient weights while preserving the overall forecasting objective and inference architecture.
    \item We evaluate BEAT on seven real-world datasets across multiple prediction horizons. The results show competitive forecasting accuracy and improvements over the WPMixer backbone, while ablation studies examine the joint effect of FSM and DGB and the choices of discrepancy measure and modulation design.
\end{itemize}

\section{Related Work}
\label{sec:related_work}

\subsection{Temporal-domain Time-Series Forecasting}
Temporal-domain methods use convolutional, linear- and MLP-based, or attention-based architectures to model dependencies across time, scales, and variables.

\paragraph{CNN-based methods} Convolutional neural networks~\cite{he2016deep,gu2018recent} learn local features through shared filters and support detection and segmentation~\cite{jiang2024weakly,li2022a3d}. For forecasting, filter range and interaction across resolutions are central design choices. SCINet~\cite{liu2022scinet} models multiple resolutions through hierarchical downsampling, convolution, and interaction. MICN~\cite{wang2023micn} combines kernels for local and global temporal modeling. ModernTCN~\cite{luo2024moderntcn} extends temporal convolutional networks~\cite{bai2018empirical} with depthwise convolution and feed-forward networks that separate temporal and cross-variable modeling.

\paragraph{Linear- and MLP-based methods} Linear models and multilayer perceptrons (MLPs) map historical observations to future values. DLinear~\cite{zeng2023transformers} applies linear models independently to each variable. Other methods use deeper residual and multiscale architectures. N-BEATS~\cite{oreshkin2019nbeats} stacks fully connected blocks in a doubly residual architecture, while TimeMixer~\cite{wang2024timemixer} mixes temporal patterns across sampling scales.

\paragraph{Transformer-based methods} Attention-based sequence modeling~\cite{vaswani2017attention} has been adapted to forecasting~\cite{zhou2021informer,zhou2022fedformer}. Autoformer~\cite{wu2021autoformer} uses auto-correlation and time-delay aggregation to combine subseries with similar periodic behavior, and Pyraformer~\cite{liu2021pyraformer} applies pyramidal attention over a multiresolution C-ary tree to capture short- and long-range dependencies with linear time and space complexity. iTransformer~\cite{liu2023itransformer} embeds each variable's history as a token and models inter-variable dependencies through attention. 

These forecasting methods primarily improve the prediction architecture to capture temporal and cross-variable dependencies. BEAT instead focuses on the training of frequency-specific predictors within a wavelet-based forecaster, using relative coefficient prediction errors to adjust their gradient weights. This complements the temporal modeling performed by each branch without changing the forward prediction architecture.

\subsection{Frequency-domain Time-Series Forecasting}
Frequency-domain methods model temporal variation through spectral components or multiresolution coefficients. Methods differ in which components they retain and how they combine information across temporal scales. TimesNet~\cite{wu2023timesnet} uses the fast Fourier transform (FFT)~\cite{brigham1988fast} to identify dominant periods and rearrange one-dimensional sequences into two-dimensional temporal representations. FFT also enables efficient series-wise correlation in Autoformer~\cite{wu2021autoformer}. FiLM~\cite{zhou2022film} combines Fourier analysis and low-rank approximation to emphasize low-frequency information, while FITS~\cite{xu2023fits} removes frequencies above a cutoff and learns from the remaining spectrum. FEDformer~\cite{zhou2022fedformer} instead randomly samples Fourier components across the spectrum. Fredformer~\cite{piao2024fredformer} addresses frequency bias associated with disproportionate attention to high-energy components. BEAT instead determines individual branch weights from current relative fitting errors, rather than assigning them according to spectral energy alone.

Other architectures also use spectral representations. EV-FGN~\cite{yi2022edge} combines a two-dimensional Fourier transform of spatiotemporal embeddings with graph convolutions. FreTS~\cite{yi2024frequency} uses MLPs to model real and imaginary spectral components, while TSLANet~\cite{eldele2024tslanet} replaces self-attention with an adaptive spectral block containing Fourier-based global and local filters for general time-series representation learning. In time-series anomaly detection, FCVAE~\cite{wang2024revisiting} conditions a variational autoencoder on global and local frequency components.

Wavelet methods preserve coefficient sequences at different temporal resolutions. WaveToken~\cite{masserano2025wavetoken} thresholds and quantizes wavelet coefficients into discrete tokens, then pre-trains an autoregressive model to forecast future coefficients. WPMixer~\cite{murad2024wpmixer} combines multilevel wavelet decomposition, patching, and mixing to predict approximation and detail coefficients with separate networks. Their outputs are reconstructed into the forecast. BEAT retains this backbone and introduces error-dependent gradient modulation during training, without replacing the frequency-specific prediction networks.

Frequency-domain supervision provides another route. FreDF~\cite{wang2025fredf} aligns predictions and labels in the frequency domain to mitigate label correlation, while Time-o1~\cite{wang2025timeo1} emphasizes significant decorrelated label components during alignment. OLMA~\cite{shi2025olma} combines Fourier- and wavelet-domain temporal supervision with a channel-domain transformation to address frequency bias and label-related uncertainty. Rather than modifying the forecasting objective, BEAT uses relative fitting errors to modulate the gradients of frequency-specific branches after back-propagation of the shared loss.

\section{Methodology}
\label{sec:method}

Table~\ref{tab:notation} summarizes the main notation used in BEAT. The detail-level index $i$ refers only to detail components, whereas the branch index $v$ covers the approximation and all detail components.

\begin{table}[!t]
\caption{Main notation used in BEAT.}
\label{tab:notation}
\centering
\footnotesize
\setlength{\tabcolsep}{4pt}
\begin{tabular}{@{}>{\raggedright\arraybackslash}p{0.27\columnwidth}>{\raggedright\arraybackslash}p{\dimexpr0.73\columnwidth-2\tabcolsep\relax}@{}}
\toprule
Symbol & Definition \\
\midrule
$T$, $K$, $N$ & Historical window length, forecasting horizon, and number of variables, respectively. \\
$X$ & Historical observations in $\mathbb{R}^{T\times N}$, from $x_{1-T}$ to $x_0$. \\
$Y$, $\hat{Y}$ & Future target and prediction in $\mathbb{R}^{K\times N}$, with target observations $y_1,\dots,y_K$. \\
$\mathcal{N}_X$ & RevIN transform using statistics from $X$ and the same affine parameters for input and target. \\
$X_{norm}$, $Y_{norm}$ & Historical input and future target transformed by $\mathcal{N}_X$, respectively. \\
$\hat{Y}_{norm}$ & Reconstructed forecast before inverse RevIN. \\
\midrule
$f$, $\psi$ & Wavelet decomposition level and wavelet type, respectively. \\
$A$, $D_i$ & Approximation component at level $f$ and detail component at level $i\in\{1,\dots,f\}$. \\
$X_A$, $X_{D_i}$ & Approximation and detail coefficient series of the normalized historical input. \\
$Y_A$, $Y_{D_i}$ & Approximation and detail coefficients of $Y_{norm}$ used as FSM targets. \\
$\hat{Y}_A$, $\hat{Y}_{D_i}$ & Predicted approximation and detail coefficients in the same normalized space as their targets. \\
$v$ & Index over the $f+1$ branches: one approximation branch and $f$ detail branches. \\
$\phi^{v}$, $\theta^{v}_{u}$ & Frequency-specific prediction network and its parameters at optimization step $u$. \\
\midrule
$\delta_A$, $\delta_{D_i}$ & MSE discrepancies of the approximation and detail coefficient predictions. \\
$\delta^{v}$ & Discrepancy of branch $v$, denoting either $\delta_A$ or a detail discrepancy $\delta_{D_i}$. \\
$\mu$ & Mean discrepancy over the $f$ detail branches only, excluding the approximation branch. \\
$r_A$, $r_{D_i}$, $r^{v}$ & Discrepancies relative to $\mu$ when $\mu>0$, with $r^{v}=\delta^{v}/\mu$ for branch $v$. \\
\midrule
$\lambda$ & Shared, tunable modulation strength, with $0<\lambda<1$. \\
$c^{v}$ & Positive gradient multiplier for branch $v$, defined in Equation~\ref{eqn:cv} for $\mu>0$. \\
$u$, $U$ & Optimization-step index and total number of training steps, respectively. \\
$\mathcal{L}_{u}$ & Forecasting loss between $\hat{Y}$ and $Y$ after inverse RevIN on the current mini-batch. \\
$g^{v}_{u}$, $\bar{g}^{v}_{u}$ & Original branch gradient $\nabla_{\theta^{v}}\mathcal{L}_{u}$ and modulated gradient $c^{v}g^{v}_{u}$. \\
$\alpha$ & Learning rate in the gradient-descent update. \\
\bottomrule
\end{tabular}
\end{table}

\subsection{Task Formulation}

Let $X \in \mathbb{R}^{T \times N}$ denote a historical window containing $T$ observations of $N$ variables. The observations are indexed from $x_{1-T}$ to $x_0$. Given this window, the task is to predict the next $K$ observations, from $y_1$ to $y_K$, represented by $Y \in \mathbb{R}^{K \times N}$. The prediction $\hat{Y}$ has the same dimensions as the target $Y$, and $K$ specifies the forecasting horizon. BEAT performs this prediction through wavelet coefficient series and uses frequency-specific discrepancies to modulate the gradients of the component networks during training.

\subsection{Overall Architecture}

\begin{figure*}[!t]
\centering
\includegraphics[width=\linewidth]{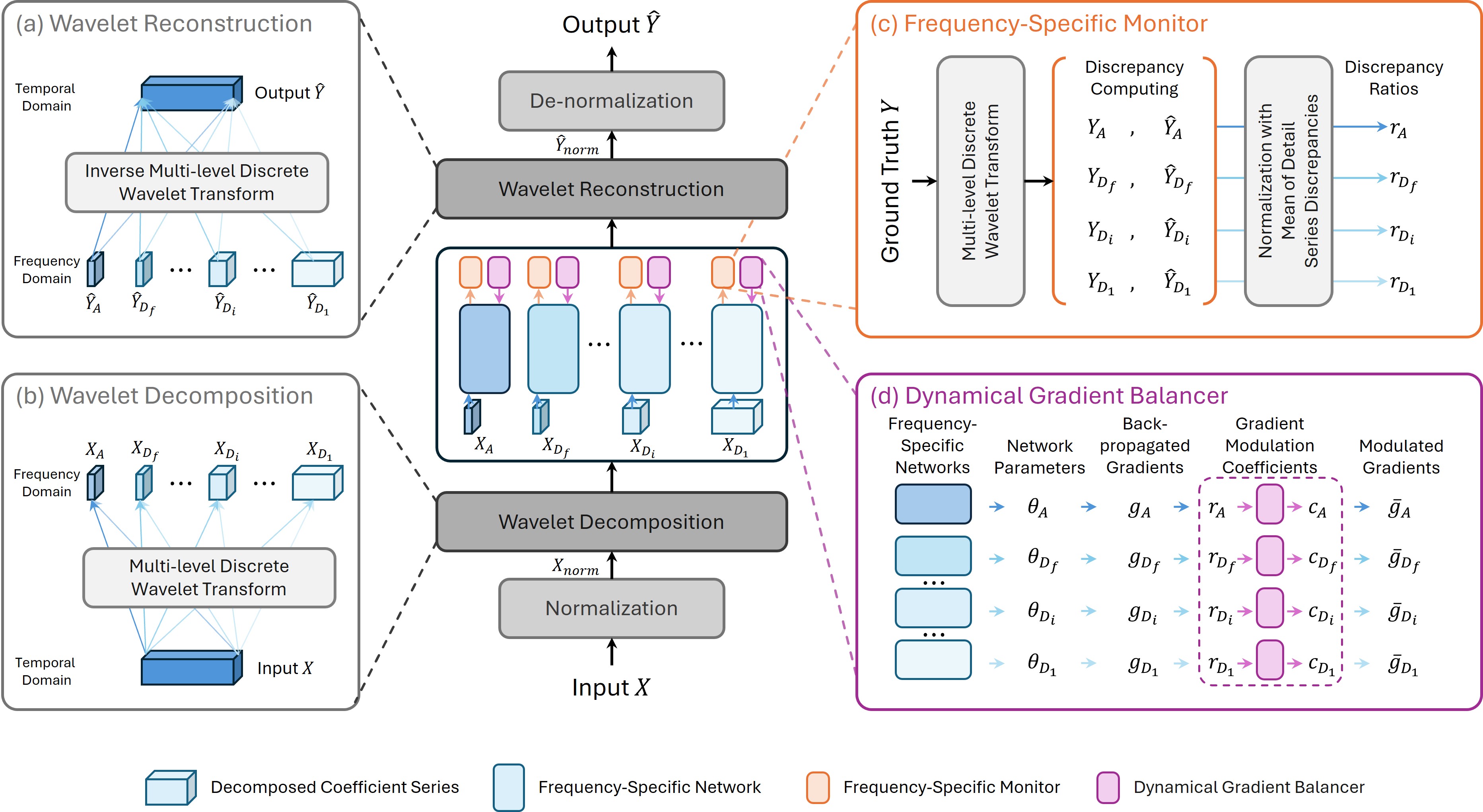}
\caption{Overall architecture of BEAT. The forecasting path decomposes historical observations into approximation and detail coefficient series, predicts their future values with frequency-specific networks, and reconstructs the forecast. (a) Wavelet reconstruction using the inverse discrete wavelet transform. (b) Multi-level wavelet decomposition. During training, (c) FSM compares predicted and target coefficients in the same normalized space to obtain relative discrepancy ratios, and (d) DGB converts these ratios into bounded coefficients that modulate the corresponding branch gradients. For clarity, panel (c) omits the target-normalization step $Y_{norm}=\mathcal{N}_X(Y)$ before wavelet decomposition, where $\mathcal{N}_X$ is the same history-conditioned RevIN transform used for the input. FSM and DGB are omitted at inference. Best viewed in color and zoomed in.}
\label{fig:overall_architecture}
\end{figure*}

Figure~\ref{fig:overall_architecture} presents the forecasting path and the training-time modulation of BEAT. We first apply Reversible Instance Normalization (RevIN)~\cite{kim2021reversible} to the historical observations $X$ to account for variations in their mean and variance. We denote this transform by $\mathcal{N}_X$, whose mean and variance are computed separately for each variable from the historical window. The resulting sequence $X_{norm}=\mathcal{N}_X(X)$ is transposed and decomposed by a multi-level discrete wavelet transform~\cite{mallat1989theory,daubechies1992ten,wang2018multilevel}. With decomposition level $f$, the retained representation consists of the approximation coefficient series $X_A$ at level $f$ and the detail coefficient series $X_{D_i}$ for $i \in \{1,2,\dots,f\}$. These $f+1$ series describe the input at different temporal resolutions. The decomposition level therefore determines the number of prediction branches, while the same monitoring and modulation principles are applied across all branches.

We instantiate this forecasting path with WPMixer~\cite{murad2024wpmixer}. Each of its $f+1$ frequency-specific networks models the temporal dependencies within one coefficient series and predicts the corresponding future coefficients. The predicted approximation $\hat{Y}_A$ and details $\hat{Y}_{D_i}$ are then passed to wavelet reconstruction, as illustrated in Figure~\ref{fig:overall_architecture}(a). The inverse transform combines these predictions into the normalized temporal-domain forecast $\hat{Y}_{norm}$. Transposition and the inverse RevIN operation restore the output organization and the scale of the forecasting target $Y$, yielding $\hat{Y}$. The inverse operation uses the statistics retained from $X$ and reverses the same affine transformation.

FSM and DGB connect this forecasting path to branch-wise training control. During training, FSM obtains coefficient targets in the same normalized space as the branch predictions and measures the discrepancy of each prediction branch, as shown in Figure~\ref{fig:overall_architecture}(c). DGB uses the resulting relative errors to scale the branch gradients before the optimizer updates their parameters, as shown in Figure~\ref{fig:overall_architecture}(d). The forecasting objective is computed between the reconstructed, inverse-normalized output $\hat{Y}$ and the target $Y$. Thus, coefficient discrepancies determine the gradient weights, while the overall forecasting loss determines the underlying gradients. These two roles allow BEAT to introduce frequency-specific modulation without changing the forward prediction architecture.

\subsection{Frequency-Specific Monitor}

We design the Frequency-Specific Monitor (FSM) to provide an error signal for each frequency-specific network. The overall forecasting loss measures the quality of the reconstructed sequence, whereas FSM measures the discrepancies of the coefficient predictions that contribute to this sequence. The design draws on balanced gradient modulation~\cite{peng2022balanced} and uses the current fitting errors to guide branch-wise gradient allocation. The predicted coefficients belong to the normalized forecasting space, so their targets must be expressed in that same space before any discrepancy is computed. FSM therefore transforms the ground-truth target $Y$ with $\mathcal{N}_X$, reusing the mean, variance, numerical stabilization, and affine parameters employed for the corresponding historical input. It does not estimate new statistics from $Y$ or replace the statistics retained for inverse RevIN. For decomposition level $f$ and wavelet type $\psi$, target construction is written as

\begin{equation}
    \begin{aligned}
        Y_{norm} &= \mathcal{N}_X(Y),\\
        (Y_A,Y_{D_f},\dots,Y_{D_1}) &= DWT(Y_{norm},f,\psi),
    \end{aligned}
    \label{eqn:dwt_y}
\end{equation}

where $DWT$ denotes the discrete wavelet transform. We retain the approximation coefficient series $Y_A$ at the final level $f$ together with all $f$ detail series $Y_{D_i}$. This organization gives each prediction branch a corresponding target at the same wavelet resolution and on the same normalization scale. Applying $DWT$ directly to the unnormalized $Y$ would instead mix scales in the subsequent comparison. The wavelet type $\psi$ specifies the transform family, and our experiments consider Coiflets, Daubechies, Biorthogonal, and Symlets.

For each branch, FSM compares the predicted coefficient series with its corresponding normalized-space target using mean squared error. The approximation and detail discrepancies are defined as

\begin{equation}
    \delta_A = MSE(Y_A, \hat{Y}_A),
    \label{eqn:mse_a}
\end{equation}

\begin{equation}
    \delta_{D_i} = MSE(Y_{D_i}, \hat{Y}_{D_i}), \quad i \in \{1, 2, \dots, f\},
    \label{eqn:mse_i}
\end{equation}

where $MSE$ averages the squared differences between the corresponding coefficient entries. Both discrepancies are nonnegative, and a smaller value indicates a closer fit to the coefficient target under this measure. To express each discrepancy relative to a common reference, we compute the mean error of the $f$ detail components

\begin{equation}
    \mu = \frac{1}{f}\sum_{i=1}^{f} \delta_{D_i}.
    \label{eqn:mu}
\end{equation}

The reference $\mu$ summarizes the current detail-prediction errors. For $\mu>0$, FSM normalizes both the approximation discrepancy and each detail discrepancy by this reference

\begin{equation}
   r_A = \frac{\delta_A}{\mu}, \quad r_{D_i} = \frac{\delta_{D_i}}{\mu}, i \in \{1, 2, \dots, f\}.
   \label{eqn:r}
\end{equation}

We use $v \in \{1,2,\dots,f+1\}$ to index the approximation branch and the $f$ detail branches, and write their discrepancies and ratios uniformly as $\delta^{v}$ and $r^{v}$. The threshold $r^{v}=1$ has a direct interpretation. Ratios above this threshold identify coefficient errors larger than $\mu$, while ratios below it identify smaller errors. The approximation branch is compared with the same reference but is excluded from its calculation, so its discrepancy does not shift the reference used for the detail branches.

FSM recomputes the reference and ratios from the current mini-batch, so a branch is assessed relative to the current detail-prediction errors. For example, a branch with an unchanged discrepancy receives a larger ratio if the detail reference decreases. This behavior makes the ratio sensitive to the relative fitting status of the components, rather than to the absolute error of one branch alone. The measurements require neither a stored loss trajectory nor an estimate of the rate of error reduction. They provide DGB with a current relative fitting-error signal, rather than a direct diagnosis of learning speed or overfitting.

\subsection{Dynamical Gradient Balancer}

We design the Dynamical Gradient Balancer (DGB) to translate the error measurements from FSM into branch-specific gradient weights. Its guiding principle is to place greater training emphasis on coefficient predictions with larger residual fitting errors, while reducing the gradient weight of those with smaller errors. Directly using the error ratio as a multiplier would allow an arbitrarily large ratio to produce an arbitrarily large coefficient. DGB instead uses a bounded mapping centered at the reference ratio of one, which provides a controlled adjustment in both directions.

Let $\theta^{v}_{u}$ denote the parameters of frequency-specific network $\phi^{v}$ at optimization step $u$, and let $\mathcal{L}_{u}$ be the overall forecasting loss on the current mini-batch. Back-propagation through the forecasting path produces the branch gradient $g^{v}_{u}=\nabla_{\theta^{v}}\mathcal{L}_{u}$. This gradient is derived from the reconstructed forecast, so it retains the connection between the coefficient predictor and the final forecasting objective. Before modulation, a gradient-descent update with learning rate $\alpha$ takes the form
\begin{equation}
    \theta^{v}_{u+1} = \theta^{v}_{u} - \alpha g^{v}_{u}.
\end{equation}

DGB assigns a scalar coefficient to this gradient using the current discrepancy ratio $r^{v}$. We use a single tunable parameter $\lambda$, shared across the frequency-specific networks, to control the modulation strength. For $0<\lambda<1$, the coefficient is defined as
\begin{equation}
    c^{v} = 1 + \lambda\frac{r^{v}-1}{r^{v}+1}.
    \label{eqn:cv}
\end{equation}

The mapping implements the intended error-dependent allocation around a neutral point. If $r^{v}>1$, then $c^{v}>1$, and DGB amplifies the gradient of a branch whose coefficient error exceeds the reference. If $0\leq r^{v}<1$, then $c^{v}<1$, reducing the gradient of a branch with a smaller error. At $r^{v}=1$, the coefficient is exactly one. The adjustment is therefore tied to a clear comparison with the reference error, and a branch can move between the amplified and attenuated cases as its relative discrepancy changes.

The coefficient increases continuously with $r^{v}$ and satisfies $1-\lambda\leq c^{v}<1+\lambda$ for finite nonnegative ratios. Its positive lower bound retains every branch's gradient direction, while the upper bound limits the amplification contributed by modulation. The parameter $\lambda$ determines the extent of this departure from the unmodulated gradient. Smaller values keep all coefficients closer to one, whereas larger values allow a stronger distinction between branches above and below the reference. The bound applies to the multiplier itself and does not assume a bound on the original gradient magnitude.

For example, a discrepancy ratio of two gives $c^{v}=1+\lambda/3$, while a ratio of one half gives $c^{v}=1-\lambda/3$. These two cases illustrate how the same rule increases or decreases the gradient weight without introducing separate branch conditions or a discontinuity at the reference. The coefficient is recomputed from the current mini-batch, so the modulation responds to changing fitting errors throughout training.

For $\mu>0$, Equation~\ref{eqn:cv} can also be written as $c^{v}=1+\lambda(\delta^{v}-\mu)/(\delta^{v}+\mu)$. This form makes the zero-reference case explicit. When $\mu=0$ but $\delta^{v}>0$, we use $c^{v}=1+\lambda$, corresponding to the upper limit of the ratio-based rule. When $\mu=\delta^{v}=0$, we set $c^{v}=1$, since neither error provides a basis for increasing or decreasing the coefficient. These definitions cover zero discrepancies without evaluating an undefined ratio.

The modulated gradient is $\bar{g}^{v}_{u}=c^{v}g^{v}_{u}$. Under gradient descent, the corresponding parameter update becomes
\begin{equation}
    \theta^{v}_{u+1} = \theta^{v}_{u} - \alpha \bar{g}^{v}_{u}.
    \label{eqn:theta_u+1}
\end{equation}

We apply $c^{v}$ to the already computed branch gradient as a fixed scalar multiplier. In particular, the coefficient calculation is not differentiated as an additional loss term, and the coefficient discrepancy does not replace $\mathcal{L}_{u}$. The modulation therefore retains the gradient direction supplied by the forecasting objective while changing its weight separately for each frequency-specific network. With other optimizers, the modulated gradient $\bar{g}^{v}_{u}$ is supplied to the optimizer's existing update rule. Equation~\ref{eqn:theta_u+1} illustrates the gradient-descent case, rather than asserting a proportional change in parameter updates for every optimizer.

\begin{algorithm}[tb]
   \caption{The proposed BEAT training framework}
   \label{alg:beat}
\begin{algorithmic}
    \STATE {\bfseries Input:} Training input--target pairs $(X,Y)$, decomposition level $f$, wavelet type $\psi$, initialized parameters $\theta^{v}$ for $v \in \{1,\dots,f+1\}$, optimizer, modulation strength $\lambda$, and total training steps $U$.
    \FOR{$u=1$ {\bfseries to} $U$}
    \STATE Sample a mini-batch of historical inputs and targets.
    \STATE Clear the gradients from the previous step.
    \STATE Normalize $X$ with $\mathcal{N}_X$, retaining its statistics.
    \STATE Predict the frequency coefficients and reconstruct $\hat{Y}_{norm}$.
    \STATE Apply inverse RevIN using the retained statistics to obtain $\hat{Y}$.
    \STATE Calculate $\mathcal{L}_{u}$ between $\hat{Y}$ and $Y$.
    \STATE Back-propagate to obtain $g^{v}_{u}=\nabla_{\theta^{v}}\mathcal{L}_{u}$.
    \STATE Form $Y_{norm}=\mathcal{N}_X(Y)$ without recomputing statistics.
    \STATE Decompose $Y_{norm}$ and compute $\delta^{v}$ and $\mu$ using Equations~\ref{eqn:dwt_y}--\ref{eqn:mu}.
    \STATE If $\mu>0$, compute $r^{v}$ using Equation~\ref{eqn:r} and $c^{v}$ using Equation~\ref{eqn:cv}.
    \STATE If $\mu=0$, set $c^{v}=1+\lambda$ for $\delta^{v}>0$ and $c^{v}=1$ otherwise.
    \STATE Replace each branch gradient with $\bar{g}^{v}_{u}=c^{v}g^{v}_{u}$.
    \STATE Update the model parameters with the optimizer.
    \ENDFOR
\end{algorithmic}
\end{algorithm}

\subsection{Training and Inference}

Algorithm~\ref{alg:beat} summarizes the training procedure. For each mini-batch, the frequency-specific networks predict the future coefficient series, which are reconstructed into the temporal-domain forecast. After inverse RevIN, the overall forecasting loss is computed between $\hat{Y}$ and $Y$ and back-propagated to obtain each branch gradient. FSM separately constructs its targets from $Y_{norm}=\mathcal{N}_X(Y)$ and measures the discrepancies between coefficient series in the shared normalized space, and DGB converts these measurements into branch-specific coefficients. The optimizer then receives the modulated gradients and updates the model parameters. We use SmoothL1Loss with a threshold of 1 as the forecasting objective.

This procedure separates error monitoring from loss optimization. The forecasting loss defines the prediction task, FSM indicates the relative fitting errors of the coefficient branches, and DGB controls how those errors influence the gradient weights. The ratios and modulation coefficients are refreshed at each training step, so the adjustment is not a fixed preference for low- or high-frequency components.

At inference, only the forward forecasting path is required. Historical observations pass through normalization, wavelet decomposition, the frequency-specific networks, wavelet reconstruction, and de-normalization to produce the forecast. Target normalization, target decomposition, and error-ratio calculation are not required. FSM and DGB therefore introduce no target-dependent operations into inference and leave the prediction architecture unchanged.

\section{Experiments}
\label{sec:experiments}

\subsection{Experimental Settings}

\paragraph{Datasets} We benchmark our method over seven challenging datasets in long-term series forecasting. 
(1) The weather dataset is collected every 10 minutes throughout 2020, containing 21 meteorological indicators such as air temperature and humidity.
(2) The ETT dataset~\cite{zhou2021informer} contains data collected from electricity transformers, including load and oil temperature measurements from July 2016 to July 2018. It comprises four subsets. ETTh1 and ETTh2 have an hourly resolution, whereas ETTm1 and ETTm2 have a 15-minute resolution.
(3) The traffic dataset comprises hourly data from the California Department of Transportation, detailing road occupancy rates measured by various sensors on San Francisco Bay Area freeways.
(4) ECL dataset~\cite{zhou2021informer} records hourly electricity consumption data from 321 clients.
Table~\ref{tab:datasets_introduction} provides data statistics of these datasets. Size denotes the number of time points in the (Train, Validation, Test) sets, respectively.

\begin{table}[t]
\centering
\begin{threeparttable}
\caption{Datasets used for evaluation.}
\label{tab:datasets_introduction}
\setlength{\tabcolsep}{4mm}{
\begin{tabular}{llll}
\toprule
Dataset      & \# VAR & Size                  & FRQ       \\
\midrule
Weather      & 21        & (36792, 5271, 10540)  & 10 min         \\
ETTh1/h2 & 7         & (8545, 2881, 2881)    & 1 hour     \\
ETTm1/m2 & 7         & (34465, 11521, 11521) & 15 min     \\
Traffic      & 862       & (12185, 1757, 3509)   & 1 hour         \\
ECL          & 321       & (18317, 2633, 5261)   & 1 hour     \\
\bottomrule
\end{tabular}}
\begin{tablenotes}
        \footnotesize
        \item[1] \# VAR denotes the number of variates.
        \item[2] FRQ means the frequency of data sampling. 
\end{tablenotes}
\end{threeparttable}
\end{table}

\paragraph{Baselines} We compare BEAT with eight forecasting baselines, namely Delayformer~\cite{wang2026delayformer}, UP-TimeMixer~\cite{zhao2026unpredictability}, WPMixer~\cite{murad2024wpmixer}, TimeMixer~\cite{wang2024timemixer}, iTransformer~\cite{liu2023itransformer}, TSMixer~\cite{chen2023tsmixer}, PatchTST~\cite{nie2022time}, and Crossformer~\cite{zhang2022crossformer}. Delayformer models delay-embedded states through a shared Transformer encoder, whereas UP-TimeMixer introduces an unpredictability-aware loss to the TimeMixer backbone. We take the horizon-specific results for these two methods from their original papers under the reported evaluation settings and calculate their Avg entries using the rule in the table caption.

\begin{table}[t]
\caption{Ablation study of designed components in BEAT. FSM and DGB represent the proposed  Frequency-Specific Monitor and Dynamical Gradient Balancer. The average performance over four prediction lengths, including $\{96, 192, 336, 720\}$, is reported for each dataset's test set.}
\label{tab:ab_beat}
\centering
\setlength{\tabcolsep}{3.2mm}{
\begin{tabular}{ccccccc}
\toprule
\multicolumn{1}{c}{\multirow{2}{*}{Index}} &
  \multicolumn{1}{c}{\multirow{2}{*}{FSM}} &
  \multicolumn{1}{c}{\multirow{2}{*}{DGB}} &
  \multicolumn{2}{c}{Weather} &
  \multicolumn{2}{c}{ETTh1} \\ \cmidrule(l){4-7} 
\multicolumn{1}{c}{} & \multicolumn{1}{c}{} & \multicolumn{1}{c}{} & MSE & MAE & MSE & MAE \\ 
\midrule
(1)                    &                      &                      &  \underline{0.243} &	\underline{0.269} &	\underline{0.422} &	\underline{0.424}    \\
(2)                    &   \checkmark                   & \checkmark           &  \textbf{0.239} &	\textbf{0.263} &	\textbf{0.415} &	\textbf{0.419} 
   \\ \bottomrule
\end{tabular}}
\end{table}

\paragraph{Implementation Details} We develop BEAT using PyTorch~\cite{paszke2019pytorch} based on the TSLib~\cite{wang2024tssurvey} codebase. We use the Adam optimizer to train BEAT. We set the input sequence length to 96 for BEAT. The prediction length $K$ varies within the range of $\{96, 192, 336, 720\}$. For each dataset, all BEAT experiments are conducted on the same training set and evaluated on the same test set. For each experimental setting, we run BEAT three times and report the mean results. A single NVIDIA RTX 4090 with 24GB memory is used for all BEAT experiments. The batch size is set to 1 during testing.

\paragraph{Metrics} We evaluate forecasting accuracy using mean squared error (MSE) and mean absolute error (MAE), averaged over all test windows, prediction steps, and variables. Following WPMixer~\cite{murad2024wpmixer}, we compute MSE and MAE on the standardized data scale. Lower values indicate better forecasting performance.

\begin{table*}[htbp]
\caption{Long-term forecasting results on seven datasets with prediction lengths $\{96, 192, 336, 720\}$. BEAT uses an input length of 96. Avg is calculated from the four displayed horizon-specific results and rounded to three decimal places. First-place counts include all 35 rows and count ties for each method. The best and second-best results are highlighted in bold red and underlined blue, respectively.}
\label{tab:compare_with_sota}
\centering
\setlength{\tabcolsep}{3.4pt}{
\begin{tabular}{c|c|cc|cc|cc|cc|cc|cc|cc|cc|cc} 
\toprule
\multicolumn{2}{c}{\multirow{2}{*}{Methods}} & \multicolumn{2}{c}{BEAT} & \multicolumn{2}{c}{UP-TimeMixer} & \multicolumn{2}{c}{Delayformer} & \multicolumn{2}{c}{WPMixer} & \multicolumn{2}{c}{TimeMixer} & \multicolumn{2}{c}{iTransformer} & \multicolumn{2}{c}{TSMixer} & \multicolumn{2}{c}{PatchTST} & \multicolumn{2}{c}{Crossformer} \\ \cmidrule(l){3-20} 
\multicolumn{2}{c}{} & \multicolumn{2}{c}{Ours} & \multicolumn{2}{c}{2026} & \multicolumn{2}{c}{2026} & \multicolumn{2}{c}{2025} & \multicolumn{2}{c}{2024} & \multicolumn{2}{c}{2024} & \multicolumn{2}{c}{2023} & \multicolumn{2}{c}{2023} & \multicolumn{2}{c}{2023} \\ \midrule
\multicolumn{2}{c|}{Metric} & \cellcolor[HTML]{F2F2F2}MSE & MAE & \cellcolor[HTML]{F2F2F2}MSE & MAE & \cellcolor[HTML]{F2F2F2}MSE & MAE & \cellcolor[HTML]{F2F2F2}MSE & MAE & \cellcolor[HTML]{F2F2F2}MSE & MAE & \cellcolor[HTML]{F2F2F2}MSE & MAE & \cellcolor[HTML]{F2F2F2}MSE & MAE & \cellcolor[HTML]{F2F2F2}MSE & MAE & \cellcolor[HTML]{F2F2F2}MSE & MAE \\ \midrule
 & 96 & \cellcolor[HTML]{F2F2F2}\bestresult{0.160} & \bestresult{0.198} & \cellcolor[HTML]{F2F2F2}0.179 & 0.232 & \cellcolor[HTML]{F2F2F2}\bestresult{0.160} & 0.210 & \cellcolor[HTML]{F2F2F2}\secondresult{0.162} & \secondresult{0.204} & \cellcolor[HTML]{F2F2F2}0.163 & 0.209 & \cellcolor[HTML]{F2F2F2}0.174 & 0.214 & \cellcolor[HTML]{F2F2F2}0.175 & 0.247 & \cellcolor[HTML]{F2F2F2}0.186 & 0.227 & \cellcolor[HTML]{F2F2F2}0.195 & 0.271 \\
 & 192 & \cellcolor[HTML]{F2F2F2}\bestresult{0.207} & \bestresult{0.241} & \cellcolor[HTML]{F2F2F2}0.221 & 0.264 & \cellcolor[HTML]{F2F2F2}\secondresult{0.208} & 0.252 & \cellcolor[HTML]{F2F2F2}0.209 & \secondresult{0.246} & \cellcolor[HTML]{F2F2F2}\secondresult{0.208} & 0.250 & \cellcolor[HTML]{F2F2F2}0.221 & 0.254 & \cellcolor[HTML]{F2F2F2}0.224 & 0.294 & \cellcolor[HTML]{F2F2F2}0.234 & 0.265 & \cellcolor[HTML]{F2F2F2}0.209 & 0.277 \\
 & 336 & \cellcolor[HTML]{F2F2F2}\secondresult{0.258} & \bestresult{0.279} & \cellcolor[HTML]{F2F2F2}0.265 & 0.304 & \cellcolor[HTML]{F2F2F2}0.265 & 0.293 & \cellcolor[HTML]{F2F2F2}0.263 & \secondresult{0.287} & \cellcolor[HTML]{F2F2F2}\bestresult{0.251} & \secondresult{0.287} & \cellcolor[HTML]{F2F2F2}0.278 & 0.296 & \cellcolor[HTML]{F2F2F2}0.262 & 0.326 & \cellcolor[HTML]{F2F2F2}0.284 & 0.301 & \cellcolor[HTML]{F2F2F2}0.273 & 0.332 \\
 & 720 & \cellcolor[HTML]{F2F2F2}\bestresult{0.329} & \bestresult{0.332} & \cellcolor[HTML]{F2F2F2}0.349 & 0.358 & \cellcolor[HTML]{F2F2F2}0.346 & 0.346 & \cellcolor[HTML]{F2F2F2}\secondresult{0.339} & \secondresult{0.338} & \cellcolor[HTML]{F2F2F2}\secondresult{0.339} & 0.341 & \cellcolor[HTML]{F2F2F2}0.358 & 0.347 & \cellcolor[HTML]{F2F2F2}0.349 & 0.348 & \cellcolor[HTML]{F2F2F2}0.356 & 0.349 & \cellcolor[HTML]{F2F2F2}0.379 & 0.401 \\ 
                                 \cmidrule{2-20}
\multirow{-5}{*}{Weather} & Avg & \cellcolor[HTML]{F2F2F2}\bestresult{0.239} & \bestresult{0.263} & \cellcolor[HTML]{F2F2F2}0.254 & 0.290 & \cellcolor[HTML]{F2F2F2}0.245 & 0.275 & \cellcolor[HTML]{F2F2F2}0.243 & \secondresult{0.269} & \cellcolor[HTML]{F2F2F2}\secondresult{0.240} & 0.272 & \cellcolor[HTML]{F2F2F2}0.258 & 0.278 & \cellcolor[HTML]{F2F2F2}0.253 & 0.304 & \cellcolor[HTML]{F2F2F2}0.265 & 0.286 & \cellcolor[HTML]{F2F2F2}0.264 & 0.320 \\
\midrule
 & 96 & \cellcolor[HTML]{F2F2F2}\bestresult{0.360} & \bestresult{0.392} & \cellcolor[HTML]{F2F2F2}0.371 & 0.395 & \cellcolor[HTML]{F2F2F2}0.379 & 0.400 & \cellcolor[HTML]{F2F2F2}\secondresult{0.368} & \secondresult{0.394} & \cellcolor[HTML]{F2F2F2}0.375 & 0.400 & \cellcolor[HTML]{F2F2F2}0.386 & 0.405 & \cellcolor[HTML]{F2F2F2}0.387 & 0.411 & \cellcolor[HTML]{F2F2F2}0.460 & 0.447 & \cellcolor[HTML]{F2F2F2}0.423 & 0.448 \\
 & 192 & \cellcolor[HTML]{F2F2F2}\bestresult{0.417} & \bestresult{0.414} & \cellcolor[HTML]{F2F2F2}0.423 & \secondresult{0.415} & \cellcolor[HTML]{F2F2F2}0.428 & 0.434 & \cellcolor[HTML]{F2F2F2}\secondresult{0.420} & 0.418 & \cellcolor[HTML]{F2F2F2}0.429 & 0.421 & \cellcolor[HTML]{F2F2F2}0.441 & 0.436 & \cellcolor[HTML]{F2F2F2}0.441 & 0.437 & \cellcolor[HTML]{F2F2F2}0.512 & 0.477 & \cellcolor[HTML]{F2F2F2}0.471 & 0.474 \\
 & 336 & \cellcolor[HTML]{F2F2F2}\bestresult{0.443} & \bestresult{0.427} & \cellcolor[HTML]{F2F2F2}0.472 & 0.449 & \cellcolor[HTML]{F2F2F2}0.476 & 0.451 & \cellcolor[HTML]{F2F2F2}\secondresult{0.452} & \secondresult{0.433} & \cellcolor[HTML]{F2F2F2}0.484 & 0.458 & \cellcolor[HTML]{F2F2F2}0.487 & 0.458 & \cellcolor[HTML]{F2F2F2}0.507 & 0.467 & \cellcolor[HTML]{F2F2F2}0.546 & 0.496 & \cellcolor[HTML]{F2F2F2}0.570 & 0.546 \\
 & 720 & \cellcolor[HTML]{F2F2F2}\bestresult{0.439} & \bestresult{0.443} & \cellcolor[HTML]{F2F2F2}0.479 & 0.465 & \cellcolor[HTML]{F2F2F2}0.468 & 0.465 & \cellcolor[HTML]{F2F2F2}\secondresult{0.449} & \secondresult{0.449} & \cellcolor[HTML]{F2F2F2}0.498 & 0.482 & \cellcolor[HTML]{F2F2F2}0.503 & 0.491 & \cellcolor[HTML]{F2F2F2}0.527 & 0.548 & \cellcolor[HTML]{F2F2F2}0.544 & 0.517 & \cellcolor[HTML]{F2F2F2}0.653 & 0.621 \\ \cmidrule{2-20}
\multirow{-5}{*}{ETTh1} & Avg & \cellcolor[HTML]{F2F2F2}\bestresult{0.415} & \bestresult{0.419} & \cellcolor[HTML]{F2F2F2}0.436 & 0.431 & \cellcolor[HTML]{F2F2F2}0.438 & 0.438 & \cellcolor[HTML]{F2F2F2}\secondresult{0.422} & \secondresult{0.424} & \cellcolor[HTML]{F2F2F2}0.447 & 0.440 & \cellcolor[HTML]{F2F2F2}0.454 & 0.448 & \cellcolor[HTML]{F2F2F2}0.466 & 0.466 & \cellcolor[HTML]{F2F2F2}0.516 & 0.484 & \cellcolor[HTML]{F2F2F2}0.529 & 0.522 \\ \midrule
 & 96 & \cellcolor[HTML]{F2F2F2}\bestresult{0.273} & \bestresult{0.329} & \cellcolor[HTML]{F2F2F2}0.285 & 0.338 & \cellcolor[HTML]{F2F2F2}0.290 & 0.338 & \cellcolor[HTML]{F2F2F2}\secondresult{0.282} & \secondresult{0.334} & \cellcolor[HTML]{F2F2F2}0.289 & 0.341 & \cellcolor[HTML]{F2F2F2}0.297 & 0.349 & \cellcolor[HTML]{F2F2F2}0.308 & 0.357 & \cellcolor[HTML]{F2F2F2}0.308 & 0.355 & \cellcolor[HTML]{F2F2F2}0.745 & 0.584 \\
 & 192 & \cellcolor[HTML]{F2F2F2}\bestresult{0.355} & \bestresult{0.378} & \cellcolor[HTML]{F2F2F2}0.365 & \secondresult{0.385} & \cellcolor[HTML]{F2F2F2}0.369 & 0.391 & \cellcolor[HTML]{F2F2F2}\secondresult{0.359} & \secondresult{0.385} & \cellcolor[HTML]{F2F2F2}0.372 & 0.392 & \cellcolor[HTML]{F2F2F2}0.380 & 0.400 & \cellcolor[HTML]{F2F2F2}0.395 & 0.404 & \cellcolor[HTML]{F2F2F2}0.393 & 0.405 & \cellcolor[HTML]{F2F2F2}0.877 & 0.656 \\
 & 336 & \cellcolor[HTML]{F2F2F2}\bestresult{0.368} & \bestresult{0.401} & \cellcolor[HTML]{F2F2F2}0.381 & 0.409 & \cellcolor[HTML]{F2F2F2}0.423 & 0.427 & \cellcolor[HTML]{F2F2F2}\secondresult{0.374} & \secondresult{0.404} & \cellcolor[HTML]{F2F2F2}0.386 & 0.414 & \cellcolor[HTML]{F2F2F2}0.428 & 0.432 & \cellcolor[HTML]{F2F2F2}0.428 & 0.434 & \cellcolor[HTML]{F2F2F2}0.427 & 0.436 & \cellcolor[HTML]{F2F2F2}1.043 & 0.731 \\
 & 720 & \cellcolor[HTML]{F2F2F2}\bestresult{0.399} & \bestresult{0.416} & \cellcolor[HTML]{F2F2F2}\secondresult{0.405} & 0.428 & \cellcolor[HTML]{F2F2F2}0.427 & 0.441 & \cellcolor[HTML]{F2F2F2}\secondresult{0.405} & \secondresult{0.427} & \cellcolor[HTML]{F2F2F2}0.412 & 0.434 & \cellcolor[HTML]{F2F2F2}0.427 & 0.445 & \cellcolor[HTML]{F2F2F2}0.443 & 0.451 & \cellcolor[HTML]{F2F2F2}0.436 & 0.450 & \cellcolor[HTML]{F2F2F2}1.104 & 0.763 \\ \cmidrule{2-20}
\multirow{-5}{*}{ETTh2} & Avg & \cellcolor[HTML]{F2F2F2}\bestresult{0.349} & \bestresult{0.381} & \cellcolor[HTML]{F2F2F2}0.359 & 0.390 & \cellcolor[HTML]{F2F2F2}0.377 & 0.399 & \cellcolor[HTML]{F2F2F2}\secondresult{0.355} & \secondresult{0.388} & \cellcolor[HTML]{F2F2F2}0.365 & 0.395 & \cellcolor[HTML]{F2F2F2}0.383 & 0.407 & \cellcolor[HTML]{F2F2F2}0.394 & 0.412 & \cellcolor[HTML]{F2F2F2}0.391 & 0.412 & \cellcolor[HTML]{F2F2F2}0.942 & 0.684 \\
\midrule
 & 96 & \cellcolor[HTML]{F2F2F2}\bestresult{0.307} & \bestresult{0.342} & \cellcolor[HTML]{F2F2F2}0.315 & 0.351 & \cellcolor[HTML]{F2F2F2}0.320 & 0.362 & \cellcolor[HTML]{F2F2F2}\secondresult{0.314} & \secondresult{0.350} & \cellcolor[HTML]{F2F2F2}0.320 & 0.357 & \cellcolor[HTML]{F2F2F2}0.334 & 0.368 & \cellcolor[HTML]{F2F2F2}0.331 & 0.378 & \cellcolor[HTML]{F2F2F2}0.352 & 0.374 & \cellcolor[HTML]{F2F2F2}0.404 & 0.426 \\
 & 192 & \cellcolor[HTML]{F2F2F2}\bestresult{0.355} & \bestresult{0.369} & \cellcolor[HTML]{F2F2F2}\bestresult{0.355} & \secondresult{0.375} & \cellcolor[HTML]{F2F2F2}\secondresult{0.356} & 0.382 & \cellcolor[HTML]{F2F2F2}0.358 & \secondresult{0.375} & \cellcolor[HTML]{F2F2F2}0.361 & 0.381 & \cellcolor[HTML]{F2F2F2}0.377 & 0.391 & \cellcolor[HTML]{F2F2F2}0.386 & 0.399 & \cellcolor[HTML]{F2F2F2}0.390 & 0.393 & \cellcolor[HTML]{F2F2F2}0.450 & 0.451 \\
 & 336 & \cellcolor[HTML]{F2F2F2}\bestresult{0.378} & \bestresult{0.390} & \cellcolor[HTML]{F2F2F2}\secondresult{0.382} & 0.398 & \cellcolor[HTML]{F2F2F2}0.388 & 0.406 & \cellcolor[HTML]{F2F2F2}0.384 & \secondresult{0.395} & \cellcolor[HTML]{F2F2F2}0.390 & 0.404 & \cellcolor[HTML]{F2F2F2}0.426 & 0.420 & \cellcolor[HTML]{F2F2F2}0.426 & 0.421 & \cellcolor[HTML]{F2F2F2}0.421 & 0.414 & \cellcolor[HTML]{F2F2F2}0.532 & 0.515 \\
 & 720 & \cellcolor[HTML]{F2F2F2}\secondresult{0.445} & \bestresult{0.427} & \cellcolor[HTML]{F2F2F2}\bestresult{0.441} & \secondresult{0.432} & \cellcolor[HTML]{F2F2F2}\secondresult{0.445} & 0.440 & \cellcolor[HTML]{F2F2F2}0.448 & \secondresult{0.432} & \cellcolor[HTML]{F2F2F2}0.454 & 0.441 & \cellcolor[HTML]{F2F2F2}0.491 & 0.459 & \cellcolor[HTML]{F2F2F2}0.489 & 0.465 & \cellcolor[HTML]{F2F2F2}0.462 & 0.449 & \cellcolor[HTML]{F2F2F2}0.666 & 0.589 \\
                                 \cmidrule{2-20}
\multirow{-5}{*}{ETTm1} & Avg & \cellcolor[HTML]{F2F2F2}\bestresult{0.371} & \bestresult{0.382} & \cellcolor[HTML]{F2F2F2}\secondresult{0.373} & 0.389 & \cellcolor[HTML]{F2F2F2}0.377 & 0.398 & \cellcolor[HTML]{F2F2F2}0.376 & \secondresult{0.388} & \cellcolor[HTML]{F2F2F2}0.381 & 0.396 & \cellcolor[HTML]{F2F2F2}0.407 & 0.410 & \cellcolor[HTML]{F2F2F2}0.408 & 0.416 & \cellcolor[HTML]{F2F2F2}0.406 & 0.408 & \cellcolor[HTML]{F2F2F2}0.513 & 0.495 \\
\midrule
 & 96 & \cellcolor[HTML]{F2F2F2}\bestresult{0.164} & \bestresult{0.245} & \cellcolor[HTML]{F2F2F2}\secondresult{0.171} & 0.255 & \cellcolor[HTML]{F2F2F2}0.175 & 0.260 & \cellcolor[HTML]{F2F2F2}\secondresult{0.171} & \secondresult{0.253} & \cellcolor[HTML]{F2F2F2}0.175 & 0.258 & \cellcolor[HTML]{F2F2F2}0.180 & 0.264 & \cellcolor[HTML]{F2F2F2}0.179 & 0.282 & \cellcolor[HTML]{F2F2F2}0.183 & 0.270 & \cellcolor[HTML]{F2F2F2}0.287 & 0.366 \\
 & 192 & \cellcolor[HTML]{F2F2F2}\bestresult{0.228} & \bestresult{0.290} & \cellcolor[HTML]{F2F2F2}\secondresult{0.232} & \secondresult{0.294} & \cellcolor[HTML]{F2F2F2}0.245 & 0.308 & \cellcolor[HTML]{F2F2F2}0.234 & \secondresult{0.294} & \cellcolor[HTML]{F2F2F2}0.237 & 0.299 & \cellcolor[HTML]{F2F2F2}0.250 & 0.309 & \cellcolor[HTML]{F2F2F2}0.244 & 0.305 & \cellcolor[HTML]{F2F2F2}0.255 & 0.314 & \cellcolor[HTML]{F2F2F2}0.414 & 0.492 \\
 & 336 & \cellcolor[HTML]{F2F2F2}\bestresult{0.285} & \bestresult{0.327} & \cellcolor[HTML]{F2F2F2}\secondresult{0.291} & 0.335 & \cellcolor[HTML]{F2F2F2}0.304 & 0.344 & \cellcolor[HTML]{F2F2F2}0.292 & \secondresult{0.333} & \cellcolor[HTML]{F2F2F2}0.298 & 0.340 & \cellcolor[HTML]{F2F2F2}0.311 & 0.348 & \cellcolor[HTML]{F2F2F2}0.320 & 0.357 & \cellcolor[HTML]{F2F2F2}0.309 & 0.347 & \cellcolor[HTML]{F2F2F2}0.597 & 0.542 \\
 & 720 & \cellcolor[HTML]{F2F2F2}\secondresult{0.384} & \bestresult{0.386} & \cellcolor[HTML]{F2F2F2}\bestresult{0.382} & \secondresult{0.390} & \cellcolor[HTML]{F2F2F2}0.401 & 0.400 & \cellcolor[HTML]{F2F2F2}0.387 & \secondresult{0.390} & \cellcolor[HTML]{F2F2F2}0.391 & 0.396 & \cellcolor[HTML]{F2F2F2}0.412 & 0.407 & \cellcolor[HTML]{F2F2F2}0.419 & 0.432 & \cellcolor[HTML]{F2F2F2}0.412 & 0.404 & \cellcolor[HTML]{F2F2F2}1.730 & 1.042 \\
                                 \cmidrule{2-20}
\multirow{-5}{*}{ETTm2} & Avg & \cellcolor[HTML]{F2F2F2}\bestresult{0.265} & \bestresult{0.312} & \cellcolor[HTML]{F2F2F2}\secondresult{0.269} & 0.319 & \cellcolor[HTML]{F2F2F2}0.281 & 0.328 & \cellcolor[HTML]{F2F2F2}0.271 & \secondresult{0.318} & \cellcolor[HTML]{F2F2F2}0.275 & 0.323 & \cellcolor[HTML]{F2F2F2}0.288 & 0.332 & \cellcolor[HTML]{F2F2F2}0.291 & 0.344 & \cellcolor[HTML]{F2F2F2}0.290 & 0.334 & \cellcolor[HTML]{F2F2F2}0.757 & 0.611 \\ \midrule
 & 96 & \cellcolor[HTML]{F2F2F2}\secondresult{0.149} & \bestresult{0.240} & \cellcolor[HTML]{F2F2F2}0.165 & 0.257 & \cellcolor[HTML]{F2F2F2}0.158 & 0.260 & \cellcolor[HTML]{F2F2F2}0.150 & \secondresult{0.241} & \cellcolor[HTML]{F2F2F2}0.153 & 0.247 & \cellcolor[HTML]{F2F2F2}\bestresult{0.148} & \bestresult{0.240} & \cellcolor[HTML]{F2F2F2}0.190 & 0.299 & \cellcolor[HTML]{F2F2F2}0.190 & 0.296 & \cellcolor[HTML]{F2F2F2}0.219 & 0.314 \\
 & 192 & \cellcolor[HTML]{F2F2F2}\bestresult{0.160} & \bestresult{0.251} & \cellcolor[HTML]{F2F2F2}0.173 & 0.263 & \cellcolor[HTML]{F2F2F2}0.172 & 0.271 & \cellcolor[HTML]{F2F2F2}\secondresult{0.162} & \secondresult{0.252} & \cellcolor[HTML]{F2F2F2}0.166 & 0.256 & \cellcolor[HTML]{F2F2F2}\secondresult{0.162} & 0.253 & \cellcolor[HTML]{F2F2F2}0.216 & 0.323 & \cellcolor[HTML]{F2F2F2}0.199 & 0.304 & \cellcolor[HTML]{F2F2F2}0.231 & 0.322 \\
 & 336 & \cellcolor[HTML]{F2F2F2}\bestresult{0.178} & \bestresult{0.268} & \cellcolor[HTML]{F2F2F2}0.182 & 0.273 & \cellcolor[HTML]{F2F2F2}0.191 & 0.289 & \cellcolor[HTML]{F2F2F2}\secondresult{0.179} & 0.270 & \cellcolor[HTML]{F2F2F2}0.185 & 0.277 & \cellcolor[HTML]{F2F2F2}\bestresult{0.178} & \secondresult{0.269} & \cellcolor[HTML]{F2F2F2}0.226 & 0.334 & \cellcolor[HTML]{F2F2F2}0.217 & 0.319 & \cellcolor[HTML]{F2F2F2}0.246 & 0.337 \\
 & 720 & \cellcolor[HTML]{F2F2F2}\bestresult{0.215} & \bestresult{0.299} & \cellcolor[HTML]{F2F2F2}0.219 & 0.308 & \cellcolor[HTML]{F2F2F2}0.230 & 0.323 & \cellcolor[HTML]{F2F2F2}\secondresult{0.217} & \secondresult{0.304} & \cellcolor[HTML]{F2F2F2}0.225 & 0.310 & \cellcolor[HTML]{F2F2F2}0.225 & 0.317 & \cellcolor[HTML]{F2F2F2}0.250 & 0.353 & \cellcolor[HTML]{F2F2F2}0.258 & 0.352 & \cellcolor[HTML]{F2F2F2}0.280 & 0.363 \\
                                 \cmidrule{2-20}
\multirow{-5}{*}{ECL} & Avg & \cellcolor[HTML]{F2F2F2}\bestresult{0.176} & \bestresult{0.265} & \cellcolor[HTML]{F2F2F2}0.185 & 0.275 & \cellcolor[HTML]{F2F2F2}0.188 & 0.286 & \cellcolor[HTML]{F2F2F2}\secondresult{0.177} & \secondresult{0.267} & \cellcolor[HTML]{F2F2F2}0.182 & 0.273 & \cellcolor[HTML]{F2F2F2}0.178 & 0.270 & \cellcolor[HTML]{F2F2F2}0.221 & 0.327 & \cellcolor[HTML]{F2F2F2}0.216 & 0.318 & \cellcolor[HTML]{F2F2F2}0.244 & 0.334 \\
\midrule
 & 96 & \cellcolor[HTML]{F2F2F2}\secondresult{0.459} & \secondresult{0.278} & \cellcolor[HTML]{F2F2F2}0.470 & 0.287 & \cellcolor[HTML]{F2F2F2}0.504 & 0.316 & \cellcolor[HTML]{F2F2F2}0.465 & 0.286 & \cellcolor[HTML]{F2F2F2}0.462 & 0.285 & \cellcolor[HTML]{F2F2F2}\bestresult{0.395} & \bestresult{0.268} & \cellcolor[HTML]{F2F2F2}0.499 & 0.344 & \cellcolor[HTML]{F2F2F2}0.526 & 0.347 & \cellcolor[HTML]{F2F2F2}0.644 & 0.429 \\
 & 192 & \cellcolor[HTML]{F2F2F2}\secondresult{0.470} & \secondresult{0.282} & \cellcolor[HTML]{F2F2F2}0.479 & 0.292 & \cellcolor[HTML]{F2F2F2}0.513 & 0.319 & \cellcolor[HTML]{F2F2F2}0.475 & 0.290 & \cellcolor[HTML]{F2F2F2}0.473 & 0.296 & \cellcolor[HTML]{F2F2F2}\bestresult{0.417} & \bestresult{0.276} & \cellcolor[HTML]{F2F2F2}0.540 & 0.370 & \cellcolor[HTML]{F2F2F2}0.522 & 0.332 & \cellcolor[HTML]{F2F2F2}0.665 & 0.431 \\
 & 336 & \cellcolor[HTML]{F2F2F2}\secondresult{0.487} & \secondresult{0.294} & \cellcolor[HTML]{F2F2F2}0.492 & \secondresult{0.294} & \cellcolor[HTML]{F2F2F2}0.510 & 0.309 & \cellcolor[HTML]{F2F2F2}0.489 & 0.296 & \cellcolor[HTML]{F2F2F2}0.498 & 0.296 & \cellcolor[HTML]{F2F2F2}\bestresult{0.433} & \bestresult{0.283} & \cellcolor[HTML]{F2F2F2}0.557 & 0.378 & \cellcolor[HTML]{F2F2F2}0.517 & 0.334 & \cellcolor[HTML]{F2F2F2}0.674 & 0.420 \\
 & 720 & \cellcolor[HTML]{F2F2F2}0.510 & 0.312 & \cellcolor[HTML]{F2F2F2}\secondresult{0.501} & \secondresult{0.309} & \cellcolor[HTML]{F2F2F2}0.547 & 0.332 & \cellcolor[HTML]{F2F2F2}0.527 & 0.318 & \cellcolor[HTML]{F2F2F2}0.506 & 0.313 & \cellcolor[HTML]{F2F2F2}\bestresult{0.467} & \bestresult{0.302} & \cellcolor[HTML]{F2F2F2}0.586 & 0.397 & \cellcolor[HTML]{F2F2F2}0.552 & 0.352 & \cellcolor[HTML]{F2F2F2}0.683 & 0.424 \\ 
                                 \cmidrule{2-20}
\multirow{-5}{*}{Traffic} & Avg & \cellcolor[HTML]{F2F2F2}\secondresult{0.482} & \secondresult{0.292} & \cellcolor[HTML]{F2F2F2}0.486 & 0.296 & \cellcolor[HTML]{F2F2F2}0.519 & 0.319 & \cellcolor[HTML]{F2F2F2}0.489 & 0.298 & \cellcolor[HTML]{F2F2F2}0.485 & 0.298 & \cellcolor[HTML]{F2F2F2}\bestresult{0.428} & \bestresult{0.282} & \cellcolor[HTML]{F2F2F2}0.546 & 0.372 & \cellcolor[HTML]{F2F2F2}0.529 & 0.341 & \cellcolor[HTML]{F2F2F2}0.667 & 0.426 \\ \midrule
\multicolumn{2}{c|}{1st Count:} & \cellcolor[HTML]{F2F2F2}\bestresult{26} & \bestresult{30} & \cellcolor[HTML]{F2F2F2}3 & 0 & \cellcolor[HTML]{F2F2F2}1 & 0 & \cellcolor[HTML]{F2F2F2}0 & 0 & \cellcolor[HTML]{F2F2F2}1 & 0 & \cellcolor[HTML]{F2F2F2}\secondresult{7} & \secondresult{6} & \cellcolor[HTML]{F2F2F2}0 & 0 & \cellcolor[HTML]{F2F2F2}0 & 0 & \cellcolor[HTML]{F2F2F2}0 & 0 \\ \bottomrule

\end{tabular}}
\end{table*}

\subsection{Long-Term Time-Series Forecasting Results}

Table~\ref{tab:compare_with_sota} compares BEAT with eight baselines on seven real-world datasets at four prediction lengths. BEAT obtains the most first-place results, with 26 for MSE and 30 for MAE across the 28 dataset--horizon settings and seven Avg rows, counting ties for each method. It achieves the lowest horizon-averaged MSE and MAE on Weather, all four ETT datasets, and ECL. On Traffic, BEAT ranks second in both average metrics, while iTransformer ranks first.

The comparison includes the 2026 methods Delayformer and UP-TimeMixer. BEAT achieves lower horizon-averaged MSE and MAE than both methods on all seven datasets under their reported evaluation settings. For example, on ETTh2, the average MSE is 0.349 for BEAT and 0.377 for Delayformer, a reduction of 7.4\%. On Weather, BEAT reduces the average MAE from UP-TimeMixer's 0.290 to 0.263, a reduction of 9.3\%. These percentages are calculated from the displayed Avg values. The advantage in average performance does not imply a lower error at every horizon. At a prediction length of 720, UP-TimeMixer obtains lower MSE on ETTm1, ETTm2, and Traffic.

BEAT has lower MSE and MAE than WPMixer in all 28 dataset--horizon settings. On ETTm2, for example, their horizon-averaged MSE/MAE values are 0.265/0.312 and 0.271/0.318, respectively. Because the forward architecture is unchanged, a plausible explanation is that error-dependent gradient weighting directs more training emphasis toward relatively poorly fitted frequency branches, complementing the representation benefits of wavelet decomposition. The lower errors after jointly enabling FSM and DGB in Table~\ref{tab:ab_beat} support this interpretation.

BEAT also achieves lower horizon-averaged MSE and MAE than TimeMixer on all seven datasets. Across Weather and the four ETT datasets, it obtains the lowest MAE at every tested prediction length. The MSE rankings vary more with the horizon, as reflected in the first-place counts.

\begin{table}[ht]
\caption{Ablation study of the wavelet type's design choice.}
\label{tab:ab_wavelet_type}
\centering
\setlength{\tabcolsep}{3mm}{
\begin{tabular}{cccccc}
\toprule
\multirow{2}{*}{Index} & \multirow{2}{*}{Wavelet Type} & \multicolumn{2}{c}{Weather}     & \multicolumn{2}{c}{ETTh1}       \\ \cmidrule(l){3-6} 
                       &                               & MSE            & MAE            & MSE            & MAE            \\ \midrule
(1)                    & Coiflets                      & 0.242          & 0.268          & \textbf{0.415} & \textbf{0.419} \\
(2)                    & Daubechies                    & \textbf{0.239} & \textbf{0.263} & \underline{0.418}    & \underline{0.420}    \\
(3)                    & Biorthogonal                  & \underline{0.240}    & \underline{0.265}    & 0.422          & \textbf{0.419} \\
(4)                    & Symlets                       & 0.244          & 0.269          & 0.419          & 0.421          \\ \bottomrule
\end{tabular}}
\end{table}

\begin{table}[ht]
\caption{Ablation study of levels of decomposition in the wavelet transform. }
\label{tab:ab_decompose_levels}
\centering
\setlength{\tabcolsep}{3mm}{
\begin{tabular}{cccccc}
\toprule
\multirow{2}{*}{Index} & \multirow{2}{*}{\makecell{Levels of \\ Decomposition $f$}} & \multicolumn{2}{c}{Weather} & \multicolumn{2}{c}{ETTh1} \\
    &   & MSE & MAE & MSE & MAE \\ \midrule
(1)                    & 1                                              & \textbf{0.239} & \underline{0.264}    & \underline{0.416}          & \textbf{0.419} \\
(2)                    & 2                                              & \textbf{0.239} & \textbf{0.263} & 0.417          & \underline{0.420}          \\
(3)                    & 3                                              & \underline{0.240}    & 0.266          & \textbf{0.415} & \textbf{0.419} \\
(4)                    & 4                                              & 0.243          & 0.265          & \underline{0.416}          & 0.421          \\
(5)                    & 5                                              & 0.244          & 0.266          & 0.418          & 0.422    \\ \bottomrule
\end{tabular}}
\end{table}

\subsection{Ablation Study}

We conduct extensive ablation studies on the test sets of the Weather and ETTh1 datasets to evaluate the effectiveness of the proposed BEAT framework. Specifically, we examine the technical components, wavelet type, decomposition level, discrepancy calculation, and gradient modulation of BEAT. For all ablation studies, the average performance over four prediction lengths, including $\{96, 192, 336, 720\}$, is reported for each dataset's test set. The best and second-best results are shown in bold and underlined, respectively.

\paragraph{Effect of BEAT} Table~\ref{tab:ab_beat} compares the model with and without the combined Frequency-Specific Monitor (FSM) and Dynamical Gradient Balancer (DGB). On Weather, the reported MSE and MAE decrease from 0.243 and 0.269 to 0.239 and 0.263, respectively. On ETTh1, the corresponding values decrease from 0.422 and 0.424 to 0.415 and 0.419. The comparison evaluates the two components together, consistent with their complementary roles in the framework. FSM supplies the relative coefficient-error measurements, and DGB uses those measurements to determine the branch gradient weights.

\paragraph{Wavelet Type} To evaluate the effect of the wavelet type, four wavelet families are considered, including Coiflets, Daubechies, Biorthogonal, and Symlets. As shown in Table~\ref{tab:ab_wavelet_type}, the best performance on Weather and ETTh1 is achieved using Daubechies and Coiflets wavelets, respectively, suggesting that the wavelet type should be chosen according to the data patterns of each dataset.

\paragraph{Levels of Decomposition} As shown in Table~\ref{tab:ab_decompose_levels}, the effect of levels $f$ for the decomposition in the wavelet transform is evaluated. The best performance of BEAT on Weather and ETTh1 datasets is achieved when the levels $f$ are set as 2 and 3, respectively. Setting the level $f$ to 4 or 5 does not improve performance. Deeper decomposition produces shorter coefficient sequences at coarser scales, which may limit temporal modeling within the corresponding prediction branches.

\begin{table}[t]
\caption{Ablation study of the design for discrepancy calculation.}
\label{tab:ab_discrepancy_calculation}
\centering
\setlength{\tabcolsep}{3.5mm}{
\begin{tabular}{cccccc}
\toprule
\multirow{2}{*}{Index} & \multirow{2}{*}{\makecell{Discrepancy \\ Calculation}} & \multicolumn{2}{c}{Weather} & \multicolumn{2}{c}{ETTh1} \\ \cmidrule(l){3-6} 
                       &                                          & MSE          & MAE          & MSE        & MAE        \\ \midrule
(1)                    & RMSE                                     & 0.244          & 0.268          & 0.421          & 0.425          \\
(2)                    & $1-R^2$                                  & 0.241          & 0.265          & 0.418          & \underline{0.423}          \\
(3)                    & MSE                                      & \textbf{0.239} & \textbf{0.263} & \textbf{0.415} & \textbf{0.419} \\
(4)                    & MAE                                      & \underline{0.240}    & \underline{0.264}    & \underline{0.416}    & \textbf{0.419} \\ \bottomrule
\end{tabular}}
\end{table}

\paragraph{Discrepancy Calculation Design} The discrepancy measure determines the signal that FSM passes to DGB. Table~\ref{tab:ab_discrepancy_calculation} compares MSE, RMSE, the R-squared-based discrepancy $1-R^2$, and MAE for measuring the difference between predicted and target coefficient series. For the R-squared variant, we use $\delta^{v}=1-R_v^2$, where $R_v^2$ is the coefficient of determination between the predicted and target coefficient series of branch $v$. Since a higher $R_v^2$ indicates a closer fit, this transformation makes larger discrepancies correspond to poorer fitting, consistent with the other error measures. The transformed discrepancies are then used to compute the detail-branch reference $\mu$ and the relative errors $r^{v}$ before applying the same DGB rule. In the reported results, MSE gives the lowest MSE and MAE on Weather and the lowest MSE on ETTh1, while tying with MAE-based discrepancy calculation for the ETTh1 MAE. The distinction between the monitoring metric and the forecasting metric is important here. The former measures coefficient discrepancies in the shared normalized space and determines the frequency-specific training signal, while the latter evaluates the reconstructed forecast after inverse RevIN. MSE weights larger coefficient residuals more heavily through squaring, providing a direct measure of residual fitting error for the relative-error calculation in FSM.

\suppressfloats[t]
\begin{table}[t]
\caption{Ablation study of the design for learning process modulation.}
\label{tab:ab_modulation}
\centering
\setlength{\tabcolsep}{3.5mm}{
\begin{tabular}{cccccc}
\toprule
\multirow{2}{*}{Index} & \multirow{2}{*}{Modulation} & \multicolumn{2}{c}{Weather} & \multicolumn{2}{c}{ETTh1} \\ \cmidrule(l){3-6} 
                       &                             & MSE          & MAE          & MSE        & MAE        \\ \midrule
(1)                    & Loss                        & \underline{0.242}          & \underline{0.268}          & \underline{0.420}          & \underline{0.421}          \\
(2)                    & Gradient                    & \textbf{0.239} & \textbf{0.263} & \textbf{0.415} & \textbf{0.419}  \\ \bottomrule
\end{tabular}}
\end{table}

\paragraph{Learning Process Modulation Design} We compare branch-specific gradient modulation with a shared-loss design that multiplies the forecasting loss by $\bar{c}=\frac{1}{f+1}\sum_{v=1}^{f+1}c^{v}$, using the coefficients from Equation~\ref{eqn:cv}. The average includes the approximation branch and all $f$ detail branches, whereas $\mu$ is computed from detail discrepancies only. We treat $\bar{c}$ as fixed during back-propagation, so it scales every branch gradient by the same value. DGB instead applies each $c^{v}$ to its corresponding branch gradient.

Table~\ref{tab:ab_modulation} shows that branch-specific gradient modulation yields lower MSE and MAE than shared-loss scaling on both Weather and ETTh1. These results are consistent with the benefit of assigning gradient weights from each branch's relative fitting error, rather than using a single multiplier for all branches.

\section{Conclusion}
\label{sec:conclusion}

We presented BEAT, a frequency-adaptive training framework for long-term time-series forecasting. The method addresses the training of wavelet-based prediction networks by connecting their current coefficient prediction errors to branch-specific gradient weights. FSM compares each predicted coefficient series with its target in a common normalized space and expresses the discrepancy relative to a reference drawn from the detail components. DGB uses these ratios to amplify gradients for branches with larger relative errors and attenuate those with smaller errors through a continuous, bounded rule.

This design retains the overall forecasting objective and the forward prediction architecture while making the gradient allocation responsive to frequency-specific fitting errors. A shared modulation-strength parameter controls the extent of the adjustment, and the positive coefficients preserve the underlying branch gradient directions. The monitoring and modulation operations are confined to training, leaving inference to the prediction networks and wavelet reconstruction. The reported evaluation on seven real-world datasets provides comparisons across multiple forecasting horizons, together with ablations of the component and modulation choices. BEAT achieves competitive forecasting performance and presents an explicit error-dependent approach to training frequency-specific predictors.

The current monitoring signal reflects coefficient discrepancies within each mini-batch and does not explicitly incorporate their evolution across training. Future work could explore history-aware monitoring to distinguish persistent fitting differences from short-term fluctuations.

\section*{Acknowledgment}
ChatGPT was used for language editing and grammar improvement.

\bibliographystyle{IEEEtran}
\bibliography{ieee_bib}

\begin{thebibliography}{10}
\providecommand{\url}[1]{#1}
\csname url@samestyle\endcsname
\providecommand{\newblock}{\relax}
\providecommand{\bibinfo}[2]{#2}
\providecommand{\BIBentrySTDinterwordspacing}{\spaceskip=0pt\relax}
\providecommand{\BIBentryALTinterwordstretchfactor}{4}
\providecommand{\BIBentryALTinterwordspacing}{\spaceskip=\fontdimen2\font plus
\BIBentryALTinterwordstretchfactor\fontdimen3\font minus
  \fontdimen4\font\relax}
\providecommand{\BIBforeignlanguage}[2]{{%
\expandafter\ifx\csname l@#1\endcsname\relax
\typeout{** WARNING: IEEEtran.bst: No hyphenation pattern has been}%
\typeout{** loaded for the language `#1'. Using the pattern for}%
\typeout{** the default language instead.}%
\else
\language=\csname l@#1\endcsname
\fi
#2}}
\providecommand{\BIBdecl}{\relax}
\BIBdecl

\bibitem{zhang2022solar}
G.~Zhang, D.~Yang, G.~Galanis, and E.~Androulakis, ``Solar forecasting with
  hourly updated numerical weather prediction,'' \emph{Renewable and
  Sustainable Energy Reviews}, vol. 154, p. 111768, 2022.

\bibitem{zhou2021informer}
H.~Zhou, S.~Zhang, J.~Peng, S.~Zhang, J.~Li, H.~Xiong, and W.~Zhang,
  ``{Informer}: Beyond efficient transformer for long sequence time-series
  forecasting,'' in
  \emph{ProceedingsoftheAAAIConferenceonArtificialIntelligence}, vol.~35,
  no.~12, 2021, pp. 11\,106--11\,115.

\bibitem{gasparin2022deep}
A.~Gasparin, S.~Lukovic, and C.~Alippi, ``Deep learning for time series
  forecasting: The electric load case,'' \emph{CAAI Transactions on
  Intelligence Technology}, vol.~7, no.~1, pp. 1--25, 2022.

\bibitem{jin2021trafficbert}
K.~Jin, J.~Wi, E.~Lee, S.~Kang, S.~Kim, and Y.~Kim, ``{TrafficBERT}:
  Pre-trained model with large-scale data for long-range traffic flow
  forecasting,'' \emph{Expert Systems with Applications}, vol. 186, p. 115738,
  2021.

\bibitem{lai2018modeling}
G.~Lai, W.-C. Chang, Y.~Yang, and H.~Liu, ``Modeling long- and short-term
  temporal patterns with deep neural networks,'' in \emph{Proceedings of the
  41st International ACM SIGIR Conference on Research and Development in
  Information Retrieval}, 2018, pp. 95--104.

\bibitem{tang2022survey}
Y.~Tang, Z.~Song, Y.~Zhu, H.~Yuan, M.~Hou, J.~Ji, C.~Tang, and J.~Li, ``A
  survey on machine learning models for financial time series forecasting,''
  \emph{Neurocomputing}, vol. 512, pp. 363--380, 2022.

\bibitem{liu2022scinet}
M.~Liu, A.~Zeng, M.~Chen, Z.~Xu, Q.~Lai, L.~Ma, and Q.~Xu, ``{SCINet}: Time
  series modeling and forecasting with sample convolution and interaction,'' in
  \emph{AdvancesinNeuralInformationProcessingSystems}, vol.~35, 2022, pp.
  5816--5828.

\bibitem{wang2023micn}
H.~Wang, J.~Peng, F.~Huang, J.~Wang, J.~Chen, and Y.~Xiao, ``{MICN}:
  Multi-scale local and global context modeling for long-term series
  forecasting,'' in \emph{InternationalConferenceonLearningRepresentations},
  2023.

\bibitem{luo2024moderntcn}
D.~Luo and X.~Wang, ``{ModernTCN}: A modern pure convolution structure for
  general time series analysis,'' in
  \emph{InternationalConferenceonLearningRepresentations}, 2024.

\bibitem{oreshkin2019nbeats}
B.~N. Oreshkin, D.~Carpov, N.~Chapados, and Y.~Bengio, ``{N-BEATS}: Neural
  basis expansion analysis for interpretable time series forecasting,'' in
  \emph{InternationalConferenceonLearningRepresentations}, 2020.

\bibitem{zeng2023transformers}
A.~Zeng, M.~Chen, L.~Zhang, and Q.~Xu, ``Are transformers effective for time
  series forecasting?'' in
  \emph{ProceedingsoftheAAAIConferenceonArtificialIntelligence}, vol.~37,
  no.~9, 2023, pp. 11\,121--11\,128.

\bibitem{wang2024timemixer}
S.~Wang, H.~Wu, X.~Shi, T.~Hu, H.~Luo, L.~Ma, J.~Y. Zhang, and J.~Zhou,
  ``{TimeMixer}: Decomposable multiscale mixing for time series forecasting,''
  in \emph{InternationalConferenceonLearningRepresentations}, 2024.

\bibitem{wu2021autoformer}
H.~Wu, J.~Xu, J.~Wang, and M.~Long, ``{Autoformer}: Decomposition transformers
  with auto-correlation for long-term series forecasting,'' in
  \emph{AdvancesinNeuralInformationProcessingSystems}, vol.~34, 2021, pp.
  22\,419--22\,430.

\bibitem{liu2021pyraformer}
S.~Liu, H.~Yu, C.~Liao, J.~Li, W.~Lin, A.~X. Liu, and S.~Dustdar,
  ``{Pyraformer}: Low-complexity pyramidal attention for long-range time series
  modeling and forecasting,'' in
  \emph{InternationalConferenceonLearningRepresentations}, 2022.

\bibitem{liu2023itransformer}
Y.~Liu, T.~Hu, H.~Zhang, H.~Wu, S.~Wang, L.~Ma, and M.~Long, ``{iTransformer}:
  Inverted transformers are effective for time series forecasting,'' in
  \emph{InternationalConferenceonLearningRepresentations}, 2024.

\bibitem{jin2024time}
M.~Jin, S.~Wang, L.~Ma, Z.~Chu, J.~Y. Zhang, X.~Shi, P.-Y. Chen, Y.~Liang,
  Y.-F. Li, S.~Pan, and Q.~Wen, ``{Time-LLM}: Time series forecasting by
  reprogramming large language models,'' in
  \emph{InternationalConferenceonLearningRepresentations}, 2024.

\bibitem{jia2024gpt4mts}
F.~Jia, K.~Wang, Y.~Zheng, D.~Cao, and Y.~Liu, ``{GPT4MTS}: Prompt-based large
  language model for multimodal time-series forecasting,'' in \emph{Proceedings
  of the AAAI Conference on Artificial Intelligence}, vol.~38, no.~21, 2024,
  pp. 23\,343--23\,351.

\bibitem{wu2023timesnet}
H.~Wu, T.~Hu, Y.~Liu, H.~Zhou, J.~Wang, and M.~Long, ``{TimesNet}: Temporal
  {2D}-variation modeling for general time series analysis,'' in
  \emph{International Conference on Learning Representations}, 2023.

\bibitem{zhou2022fedformer}
T.~Zhou, Z.~Ma, Q.~Wen, X.~Wang, L.~Sun, and R.~Jin, ``{FEDformer}: Frequency
  enhanced decomposed transformer for long-term series forecasting,'' in
  \emph{InternationalConferenceonMachineLearning}, ser. Proceedings of Machine
  Learning Research, vol. 162.\hskip 1em plus 0.5em minus 0.4em\relax PMLR,
  2022, pp. 27\,268--27\,286.

\bibitem{piao2024fredformer}
X.~Piao, Z.~Chen, T.~Murayama, Y.~Matsubara, and Y.~Sakurai, ``{Fredformer}:
  Frequency debiased transformer for time series forecasting,'' in
  \emph{Proceedings of the 30th ACM SIGKDD Conference on Knowledge Discovery
  and Data Mining}, 2024, pp. 2400--2410.

\bibitem{masserano2025wavetoken}
L.~Masserano, A.~F. Ansari, B.~Han, X.~Zhang, C.~Faloutsos, M.~W. Mahoney,
  A.~G. Wilson, Y.~Park, S.~S. Rangapuram, D.~C. Maddix, and B.~Wang,
  ``Enhancing foundation models for time series forecasting via wavelet-based
  tokenization,'' in \emph{Proceedings of the 42nd International Conference on
  Machine Learning}, ser. Proceedings of Machine Learning Research, vol.
  267.\hskip 1em plus 0.5em minus 0.4em\relax PMLR, 2025, pp. 43\,248--43\,275.

\bibitem{murad2024wpmixer}
M.~M.~N. Murad, M.~Aktukmak, and Y.~Yilmaz, ``{WPMixer}: Efficient
  multi-resolution mixing for long-term time series forecasting,'' in
  \emph{ProceedingsoftheAAAIConferenceonArtificialIntelligence}, vol.~39,
  no.~18, 2025, pp. 19\,581--19\,588.

\bibitem{wang2025fredf}
H.~Wang, L.~Pan, Y.~Shen, Z.~Chen, D.~Yang, Y.~Yang, S.~Zhang, X.~Liu, H.~Li,
  and D.~Tao, ``{FreDF}: Learning to forecast in the frequency domain,'' in
  \emph{InternationalConferenceonLearningRepresentations}, 2025, pp.
  6893--6922.

\bibitem{wang2025timeo1}
H.~Wang, L.~Pan, Z.~Chen, X.~Chen, Q.~Dai, L.~Wang, H.~Li, and Z.~Lin,
  ``{Time-o1}: Time-series forecasting needs transformed label alignment,'' in
  \emph{AdvancesinNeuralInformationProcessingSystems}, vol.~38, 2025, pp.
  8660--8690.

\bibitem{shi2025olma}
T.~Shi, Z.~Meng, Y.~Chen, S.~Zheng, F.~Su, J.~Huang, C.~Ren, and Z.~Zhao,
  ``{OLMA}: One loss for more accurate time series forecasting,'' \emph{arXiv
  preprint arXiv:2505.11567}, 2025.

\bibitem{peng2022balanced}
X.~Peng, Y.~Wei, A.~Deng, D.~Wang, and D.~Hu, ``Balanced multimodal learning
  via on-the-fly gradient modulation,'' in
  \emph{ProceedingsoftheIEEE/CVFConferenceonComputerVisionandPatternRecognition},
  2022, pp. 8238--8247.

\bibitem{he2016deep}
K.~He, X.~Zhang, S.~Ren, and J.~Sun, ``Deep residual learning for image
  recognition,'' in \emph{Proceedings of the IEEE conference on computer vision
  and pattern recognition}, 2016, pp. 770--778.

\bibitem{gu2018recent}
J.~Gu, Z.~Wang, J.~Kuen, L.~Ma, A.~Shahroudy, B.~Shuai, T.~Liu, X.~Wang,
  G.~Wang, J.~Cai, and T.~Chen, ``Recent advances in convolutional neural
  networks,'' \emph{Pattern Recognition}, vol.~77, pp. 354--377, 2018.

\bibitem{jiang2024weakly}
X.~Jiang, S.~Jin, L.~Lu, X.~Zhang, and S.~Lu, ``Weakly supervised monocular
  3{D} detection with a single-view image,'' in
  \emph{ProceedingsoftheIEEE/CVFConferenceonComputerVisionandPatternRecognition},
  2024, pp. 10\,508--10\,518.

\bibitem{li2022a3d}
Z.~Li, W.~Ye, T.~Jiang, and T.~Huang, ``2{D} amodal instance segmentation
  guided by 3{D} shape prior,'' in \emph{ComputerVision--ECCV2022}, 2022, pp.
  165--181.

\bibitem{bai2018empirical}
S.~Bai, J.~Z. Kolter, and V.~Koltun, ``An empirical evaluation of generic
  convolutional and recurrent networks for sequence modeling,'' \emph{arXiv
  preprint arXiv:1803.01271}, 2018.

\bibitem{vaswani2017attention}
A.~Vaswani, N.~Shazeer, N.~Parmar, J.~Uszkoreit, L.~Jones, A.~N. Gomez,
  {\L}.~Kaiser, and I.~Polosukhin, ``Attention is all you need,'' in
  \emph{AdvancesinNeuralInformationProcessingSystems}, vol.~30, 2017, pp.
  5998--6008.

\bibitem{brigham1988fast}
E.~O. Brigham, \emph{The Fast {Fourier} Transform and Its Applications}.\hskip
  1em plus 0.5em minus 0.4em\relax Prentice Hall, 1988.

\bibitem{zhou2022film}
T.~Zhou, Z.~Ma, X.~Wang, Q.~Wen, L.~Sun, T.~Yao, W.~Yin, and R.~Jin, ``{FiLM}:
  Frequency improved {Legendre} memory model for long-term time series
  forecasting,'' in \emph{AdvancesinNeuralInformationProcessingSystems},
  vol.~35, 2022, pp. 12\,677--12\,690.

\bibitem{xu2023fits}
Z.~Xu, A.~Zeng, and Q.~Xu, ``{FITS}: Modeling time series with $10k$
  parameters,'' in \emph{InternationalConferenceonLearningRepresentations},
  2024.

\bibitem{yi2022edge}
K.~Yi, Q.~Zhang, L.~Hu, H.~He, N.~An, L.~Cao, and Z.~Niu, ``Edge-varying
  {Fourier} graph networks for multivariate time series forecasting,''
  \emph{arXiv preprint arXiv:2210.03093}, 2022.

\bibitem{yi2024frequency}
K.~Yi, Q.~Zhang, W.~Fan, S.~Wang, P.~Wang, H.~He, N.~An, D.~Lian, L.~Cao, and
  Z.~Niu, ``Frequency-domain {MLPs} are more effective learners in time series
  forecasting,'' in \emph{AdvancesinNeuralInformationProcessingSystems},
  vol.~36, 2023, pp. 76\,656--76\,679.

\bibitem{eldele2024tslanet}
E.~Eldele, M.~Ragab, Z.~Chen, M.~Wu, and X.~Li, ``{TSLANet}: Rethinking
  transformers for time series representation learning,'' in \emph{Proceedings
  of the 41st International Conference on Machine Learning}, ser. Proceedings
  of Machine Learning Research, vol. 235.\hskip 1em plus 0.5em minus
  0.4em\relax PMLR, 2024, pp. 12\,409--12\,428.

\bibitem{wang2024revisiting}
Z.~Wang, C.~Pei, M.~Ma, X.~Wang, Z.~Li, D.~Pei, S.~Rajmohan, D.~Zhang, Q.~Lin,
  H.~Zhang, J.~Li, and G.~Xie, ``Revisiting {VAE} for unsupervised time series
  anomaly detection: A frequency perspective,'' in \emph{Proceedings of the ACM
  Web Conference 2024}, 2024, pp. 3096--3105.

\bibitem{kim2021reversible}
T.~Kim, J.~Kim, Y.~Tae, C.~Park, J.-H. Choi, and J.~Choo, ``Reversible instance
  normalization for accurate time-series forecasting against distribution
  shift,'' in \emph{InternationalConferenceonLearningRepresentations}, 2022.

\bibitem{mallat1989theory}
S.~G. Mallat, ``A theory for multiresolution signal decomposition: the wavelet
  representation,''
  \emph{IEEETransactionsonPatternAnalysisandMachineIntelligence}, vol.~11,
  no.~7, pp. 674--693, 1989.

\bibitem{daubechies1992ten}
I.~Daubechies, \emph{Ten lectures on wavelets}, ser. CBMS-NSF Regional
  Conference Series in Applied Mathematics.\hskip 1em plus 0.5em minus
  0.4em\relax Society for Industrial and Applied Mathematics, 1992, vol.~61.

\bibitem{wang2018multilevel}
J.~Wang, Z.~Wang, J.~Li, and J.~Wu, ``Multilevel wavelet decomposition network
  for interpretable time series analysis,'' in \emph{Proceedings of the 24th
  ACM SIGKDD International Conference on Knowledge Discovery \& Data Mining},
  2018, pp. 2437--2446.

\bibitem{wang2026delayformer}
Z.~Wang, P.~Tao, and L.~Chen, ``{Delayformer}: Spatiotemporal transformation
  for predicting high-dimensional dynamics,'' \emph{Advanced Science}, vol.~13,
  no.~3, p. e14671, 2026.

\bibitem{zhao2026unpredictability}
L.~Zhao, J.~Dong, R.~Chen, Y.~Wang, Y.~Jin, and Y.~Zhao, ``Perceiving
  unpredictability for new energy power and electricity consumption
  forecasting,'' \emph{Entropy}, vol.~28, no.~1, p.~64, 2026.

\bibitem{chen2023tsmixer}
\BIBentryALTinterwordspacing
S.-A. Chen, C.-L. Li, N.~C. Yoder, S.~{\"O}. Ar{\i}k, and T.~Pfister,
  ``{TSMixer}: An all-{MLP} architecture for time series forecasting,''
  \emph{Transactions on Machine Learning Research}, 2023. [Online]. Available:
  \url{https://arxiv.org/abs/2303.06053}
\BIBentrySTDinterwordspacing

\bibitem{nie2022time}
Y.~Nie, N.~H. Nguyen, P.~Sinthong, and J.~Kalagnanam, ``A time series is worth
  64 words: Long-term forecasting with transformers,'' in
  \emph{InternationalConferenceonLearningRepresentations}, 2023.

\bibitem{zhang2022crossformer}
Y.~Zhang and J.~Yan, ``{Crossformer}: Transformer utilizing cross-dimension
  dependency for multivariate time series forecasting,'' in
  \emph{InternationalConferenceonLearningRepresentations}, 2023.

\bibitem{paszke2019pytorch}
A.~Paszke, S.~Gross, F.~Massa, A.~Lerer, J.~Bradbury, G.~Chanan, T.~Killeen,
  Z.~Lin, N.~Gimelshein, L.~Antiga, A.~Desmaison, A.~Kopf, E.~Yang, Z.~DeVito,
  M.~Raison, A.~Tejani, S.~Chilamkurthy, B.~Steiner, L.~Fang, J.~Bai, and
  S.~Chintala, ``{PyTorch}: An imperative style, high-performance deep learning
  library,'' in \emph{AdvancesinNeuralInformationProcessingSystems}, vol.~32,
  2019, pp. 8024--8035.

\bibitem{wang2024tssurvey}
Y.~Wang, H.~Wu, J.~Dong, Y.~Liu, C.~Wang, M.~Long, and J.~Wang, ``Deep time
  series models: A comprehensive survey and benchmark,''
  \emph{IEEETransactionsonPatternAnalysisandMachineIntelligence}, vol.~48,
  no.~10, pp. 11\,642--11\,662, 2026.

\end{thebibliography}

\end{document}